%% file: main.tex
\documentclass{article}

\usepackage[utf8]{inputenc} 
\usepackage[T1]{fontenc}    
\usepackage{hyperref}       
\usepackage{url}            
\usepackage{booktabs}       
\usepackage{amsfonts}       
\usepackage{nicefrac}       
\usepackage{microtype}      
\usepackage{comment}
\usepackage{mathtools}
\usepackage{xcolor}
\usepackage{wrapfig}
\usepackage{bbm}
\usepackage{algorithm} 
\usepackage{algorithmic}  
\usepackage[algo2e,linesnumbered,ruled]{algorithm2e} 
\usepackage{thm-restate}
\usepackage{capt-of}
\usepackage{lineno}
\usepackage{changepage}
\usepackage{dblfloatfix}
\usepackage{textcomp}

\input{format/cmd.tex}

\let\oldsum\sum
\renewcommand{\sum}{\textstyle\oldsum}
\newcommand{\Tau}{\mathrm{T}}
\newcommand{\minus}{\scalebox{0.5}[1.0]{$-$}}

\usepackage{enumitem,kantlipsum}
\usepackage{xcolor,pifont}%
\newcommand*\colourcheck[1]{%
  \expandafter\newcommand\csname #1check\endcsname{\textcolor{#1}{\ding{51}}}%
}
\newcommand*\colourx[1]{%
  \expandafter\newcommand\csname #1x\endcsname{\textcolor{#1}{\ding{55}}}%
}
\colourcheck{green}
\colourx{red}

\usepackage[accepted]{format/icml2020}

\icmltitlerunning{Domain Adaptive Imitation Learning}

\begin{document}

\twocolumn[
\icmltitle{Domain Adaptive Imitation Learning}



\icmlsetsymbol{equal}{*}

\begin{icmlauthorlist}
\icmlauthor{Kuno Kim}{sc}
\icmlauthor{Yihong Gu}{th}
\icmlauthor{Jiaming Song}{sc}
\icmlauthor{Shengjia Zhao}{sc}
\icmlauthor{Stefano Ermon}{sc}
\end{icmlauthorlist}

\icmlaffiliation{sc}{Department of Computer Science, Stanford University}
\icmlaffiliation{th}{Department of Computer Science, Tsinghua University}


\icmlcorrespondingauthor{Kuno Kim}{khkim@cs.stanford.edu}

\icmlkeywords{Machine Learning, ICML}

\vskip 0.3in
]



\printAffiliationsAndNotice{}  

\input{sections/abstract}
\input{sections/introduction}

\input{sections/setup}

\input{sections/alignability}

\input{sections/algorithm}

\begin{table*}[h!]
\centering
\captionof{table}{\textbf{MDP Alignment Performance}. Mean $\ell_{2}$ loss between the learned state map predictions and the ground truth permutation. On average, GAMA has $17.3\times$ lower loss than the best baseline. Results are averaged across 5 seeds.}
\vspace{5pt}
\begin{small}
\begin{sc}
\begin{tabular}{ ccccccc } 
 \toprule
 & \textbf{GAMA (Ours)} & CCA & UMA  & IF & IfO & Random \\ 
 \midrule
 pen $\leftrightarrow$ pen & $\textbf{0.057} \pm \textbf{0.017}$ & $0.72 \pm 0.25$ & >$100$ & $2.50 \pm 1.08$ & $2.24 \pm 0.82$ & >$100$ \\ 
 pen $\leftrightarrow$ cart & $\textbf{0.178} \pm \textbf{0.051}$ & $3.92 \pm 3.77$ & >$100$ & $1.62 \pm 0.52$ & $3.31 \pm 1.2$ & >$100$\\
 reach2$\leftrightarrow$reach2-tp & $\textbf{0.092} \pm \textbf{0.043}$ & $10.14 \pm 5.31$ & >$100$ & $12.41 \pm 3.12$ & $5.12 \pm 2.41$ & >$100$\\ 
 \bottomrule
\end{tabular}
\end{sc}
\end{small}
\label{table:identity}
\vspace{-10pt}
\end{table*}
\input{sections/relatedworks}

\input{sections/experiments}

\input{sections/conclusion}
\input{sections/acknowledgements}

\bibliographystyle{format/icml2020}
\bibliography{main}

\newpage

\onecolumn
\begin{center}
\Large \textbf{Domain Adaptive Imitation Learning - Supplementary Materials}
\end{center}

\appendix
\input{sections/supplementary.tex}

\end{document}

%% file: format/cmd.tex
\usepackage{amsmath,amsfonts,bm,amssymb}
\usepackage{amsthm}

\newtheorem{definition}{Definition}

\newtheorem{lemma}{Lemma}
\newtheorem{corollary}{Corollary}

\usepackage[makeroom]{cancel}

\usepackage{mathtools,stackengine}
\stackMath
\newcommand{\stackEq}[1]{%
  \setbox0=\hbox{${}\mathrel{\stackon[-1pt]{=}{\scriptstyle\text{#1\strut}}}{}$}
  \xdef\tmpwd{\dimexpr\the\wd0\relax}
  \kern.5\tmpwd\mathclap{\box0}&\kern.5\tmpwd
}

\def\eqref#1{equation~\ref{#1}}

\def\1{\bm{1}}

\DeclareMathAlphabet{\mathsfit}{\encodingdefault}{\sfdefault}{m}{sl}
\SetMathAlphabet{\mathsfit}{bold}{\encodingdefault}{\sfdefault}{bx}{n}

\def\gA{{\mathcal{A}}}

\def\gM{{\mathcal{M}}}

\def\gP{{\mathcal{P}}}

\def\gS{{\mathcal{S}}}

\def\sR{{\mathbb{R}}}

\newcommand{\E}{\mathbb{E}}

\DeclareMathOperator*{\argmax}{arg\,max}

\newcommand{\mc}[1]{{\mathcal{#1}}}
\newcommand{\bb}[1]{{\mathbb{#1}}}

\newcommand{\sx}{{{\mathcal{S}_x}}}
\newcommand{\sy}{{{\mathcal{S}_y}}}
\newcommand{\ax}{{{\mathcal{A}_x}}}
\newcommand{\ay}{{{\mathcal{A}_y}}}
\newcommand{\px}{{{P_x}}}
\newcommand{\py}{{{P_y}}}
\newcommand{\rx}{{{R_x}}}
\newcommand{\ry}{{{R_y}}}
\newcommand{\mx}{{{\mathcal{M}_x}}}
\newcommand{\my}{{{\mathcal{M}_y}}}

\renewcommand{\S}{\mathcal{S}}
\newcommand{\A}{\mathcal{A}}
\newcommand{\T}{\mathcal{T}}

\newcommand{\sigxx}{{\sigma_{\hat{\pi}_x}^x}}
\newcommand{\sigxy}{{\sigma_{\hat{\pi}_x}^{x \to y}}}
\newcommand{\sigyy}{{\sigma_{\pi_y}^y}}

\newcommand{\rhoxx}{{\rho_{\hat{\pi}_x}^x}}

%% file: sections/abstract.tex
\begin{abstract}
We study the question of how to imitate tasks across domains with discrepancies such as embodiment, viewpoint, and dynamics mismatch. Many prior works require paired, aligned demonstrations and an additional RL step that requires environment interactions. However, paired, aligned demonstrations are seldom obtainable and RL procedures are expensive. 
We formalize the Domain Adaptive Imitation Learning (DAIL) problem, which is a unified framework for imitation learning in the presence of viewpoint, embodiment, and dynamics mismatch. Informally, DAIL is the process of learning how to perform a task optimally, given demonstrations of the task in a distinct domain. We propose a two step approach to DAIL: alignment followed by adaptation. In the alignment step we execute a novel unsupervised MDP alignment algorithm, Generative Adversarial MDP Alignment (GAMA), to learn state and action correspondences from \emph{unpaired, unaligned} demonstrations. In the adaptation step we leverage the correspondences to zero-shot imitate tasks across domains. To describe when DAIL is feasible via alignment and adaptation, we introduce a theory of MDP alignability.
We experimentally evaluate GAMA against baselines in embodiment, viewpoint, and dynamics mismatch scenarios where aligned demonstrations don't exist and show the effectiveness of our approach.
\end{abstract}

%% file: sections/introduction.tex
\begin{figure*}[t]
    \centering
    \includegraphics[width=1\textwidth]{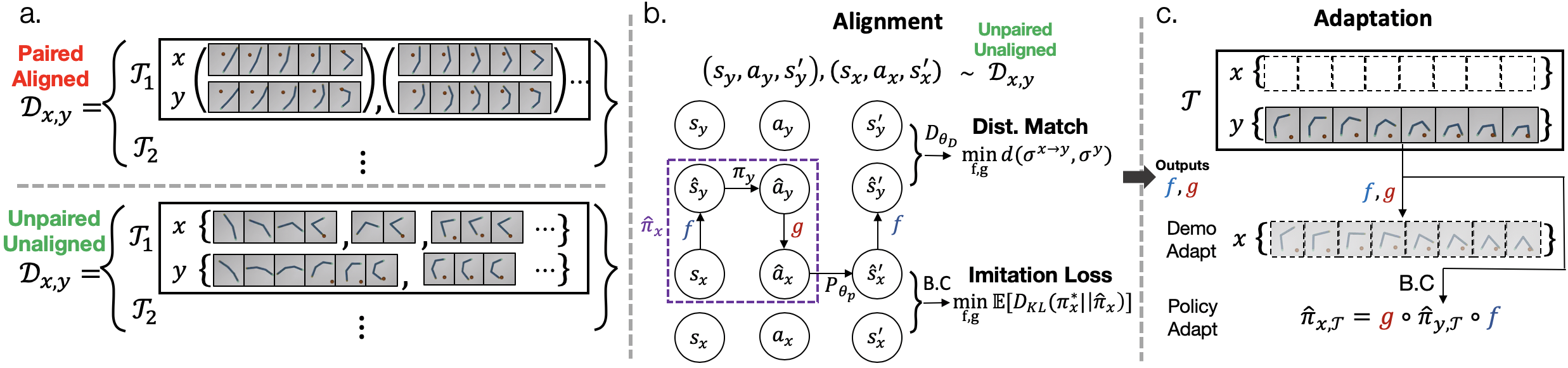}
    \vspace{-20pt}
    \caption{\textbf{DAIL Pipeline}. (a). \emph{Inputs}: Illustration of paired, aligned vs unpaired, unaligned demonstrations in the alignment task set $\mc{D}_{x, y}$ containing tasks $T_{1}, T_{2}, ...$ (b). \emph{Alignment} phase: we learn state, action maps $f, g$ between the self ($x$) and expert ($y$) domain from unpaired, unaligned demonstrations by minimizing a distribution matching loss and an imitation loss on a composite policy $\hat{\pi}_{x}$ (c) \emph{Adaptation} phase: adapt the expert domain policy $\pi_{y, \T}$ or demonstrations to obtain a self domain policy $\hat{\pi}_{x, \T}$}
    \label{fig:setup}
    \vspace{-10pt}
\end{figure*}

\section{Introduction} 
Humans possess an astonishing ability to recognize latent structural similarities between behaviors in related but distinct domains, and learn new skills from cross domain demonstrations alone. Not only are we capable of learning from third person observations that have no obvious correspondence to our internal self representations~\citep{stadie2017third, liu2018fromobs, sermanet2018tcn}, but we also are capable of imitating experts with different embodiments ~\citep{Gupta2017Learning, Rizzolatti2004Mirror, liu2020statealignment} in foreign environments~\citep{liu2020statealignment} - e.g an infant is able to imitate visuomotor skills by watching adults with different biomechanics~\citep{Jones2009Infancy} acting in environments distinct from their playroom. Previous work in neuroscience~\citep{Marshall2015Bodymap} and robotics~\citep{Kuniyoshi1993Actseq, Kuniyoshi1994Extracting} have recognized the pitfalls of exact behavioral cloning in the presence of domain discrepancies and posited that the effectiveness of the human imitation learning mechanism hinges on the ability to learn structure preserving domain correspondences. These correspondences enable the learner to internalize cross domain demonstrations and produce a reconstruction of the behavior in the self domain. Consider a young child that has learned to associate (or "align") his internal body map with the limbs of an adult. When the adult demonstrates running, the child is able internalize the demonstration, and reproduce the behavior. 

Recently, separate solutions have been proposed for imitation learning across three main types of domain discrepancies: dynamics \citep{liu2020statealignment}, embodiment \citep{Gupta2017Learning}, and viewpoint \citep{liu2018fromobs, sermanet2018tcn} mismatch. Most works \citep{liu2018fromobs, sermanet2018tcn, Gupta2017Learning} require \emph{paired, time-aligned demonstrations} to obtain state correspondences and an RL step involving environment interactions. (see Figure \ref{fig:setup}a) Unfortunately, paired, aligned demonstrations are seldom obtainable and RL procedures are expensive. 

In this work we formalize the Domain Adaptive Imitation Learning (DAIL) problem - a unified framework for imitation learning across domains with dynamics, embodiment, and/or viewpoint mismatch. Informally, DAIL is the \emph{process of learning how to perform a task optimally in a self domain, given demonstrations of the task in a distinct expert domain}. We propose a two-step approach to DAIL: alignment followed by adaptation. In the alignment step we execute a novel unsupervised MDP alignment algorithm, Generative Adversarial MDP Alignment (GAMA), to learn state, action maps from \emph{unpaired, unaligned} demonstrations. In the adaptation step we leverage the learned state, action maps to zero-shot imitate tasks across domains without an extra RL step. To shed light on when DAIL can be solved by alignment and adaptation, we introduce a theory of MDP alignability. We conduct experiments with a variety of domains that have dynamics, embodiment, and viewpoint mismatch and demonstrate significant gains on learning from unpaired data. The primary contributions of this work are as follows:
\vspace{-5pt}
\begin{enumerate}[wide, labelwidth=!, labelindent=0pt]
\item We propose an unsupervised MDP alignment algorithm that succeeds at DAIL from \emph{unpaired, unaligned demonstrations} removing the need for costly paired, aligned data.

\item We achieve \emph{zero-shot imitation}, thereby removing a costly RL procedure involving environment interactions.

\item We propose a \emph{unifying theoretical framework} for imitation learning across domains with dynamics, embodiment, and/or viewpoint mismatch. 
\end{enumerate}



%% file: sections/setup.tex
\section{Domain Adaptive Imitation Learning}
\label{section:setup}
An infinite horizon Markov Decision Process (MDP) $\mc{M} \in \Omega$ with deterministic dynamics is a tuple $(\S, \A, P, \eta, R)$ where $\Omega$ is the set of all MDPs, $\S$ is the state space, $\A$ is the action space, $P: \S \times \A \rightarrow \S$ is a (deterministic) transition function, $R: \S \times \A \rightarrow \mathbb{R}$ is the reward function, and $\eta$ is the initial state distribution. A domain is an MDP without the reward, i.e $(\S, \A, P, \eta)$. Intuitively, a domain fully characterizes the embodied agent and the environment dynamics, but not the desired behavior. A task $\T$ is a label for an MDP corresponding to the high level description of optimal behavior, such as "walking". 
$\T$ is analogous to category labels for images. 
An MDP with domain $x$ for task $\T$ is denoted by $\mx_{, \T} = (\sx, \ax, \px, \eta_{x}, \rx_{, \T})$, where $\rx_{, \T}$ is a reward function encapsulating the behavior labeled by $\T$. For example, different reward functions are needed to realize the "walking" behavior in two morphologically different humanoids. A (stationary) policy for $\mx_{, \T}$ is a map $\pi_{x, \T}: \S_{x} \rightarrow \mc{B}(\mc{A}_{x})$ where $\mc{B}$ is the set of probability measures on $\ax$ and an optimal policy $\pi_{x, \T}^* = \argmax_{\pi_{x}} J(\pi_{x})$ achieves the highest policy performance $J(\pi_{x}) = \E_{\pi_{x}}[\sum_{t = 0}^{\infty} \gamma^{t} \rx_{, \T}(s_{x}^{(t)}, a_{x}^{(t)})]$ where $0 < \gamma < 1$ is a discount factor. A demonstration $\tau_{\mx_{, \T}}$ of length $H$ for an MDP $\mx_{, \T}$ is a sequence of state, action tuples $\tau_{\mx_{, \T}} = \{(s_{x}^{(t)}, a_{x}^{(t)})\}_{t = 1}^{H}$ and $\mc{D}_{\mx_{, \T}} = \{\tau^{(k)}_{\mx_{, \T}}\}_{k = 1}^{K}$ is a set of demonstrations.

Let $\mx_{, \T}, \my_{, \T}$ be self and expert MDPs for a target task $\T$. Given expert domain demonstrations $\mc{D}_{\my_{, \T}}$, Domain Adaptive Imitation Learning (DAIL) aims to determine an optimal self domain policy $\pi^*_{x, \T}$ without access to the reward function $\rx_{, \T}$.
In this work we propose to first solve an MDP alignment problem and then leverage the alignments to zero-shot imitate expert domain demonstrations. Like prior work \citep{Gupta2017Learning, liu2018fromobs, sermanet2018tcn}, we assume the availability of an alignment task set $\mc{D}_{x, y} = \{(\mc{D}_{\mx_{, \T_{i}}}, \mc{D}_{\my_{, \T_{i}}})\}_{i = 1}^{N}$ containing demonstrations for $N$ tasks $\{\T_{i}\}_{i = 1}^{N}$ from both the self and expert domain. $\mc{D}_{x, y}$ could, for example, contain both robot ($x$) and human ($y$) demonstrations for a set primitive tasks such as walking, running, and jumping. Unlike prior work, demonstrations are \emph{unpaired and unaligned}, i.e $(s_{x}^{(t)}, s_{y}^{(t)})$ may not be a valid state correspondence (see Figure \ref{fig:setup}(a)). 
Paired, time-aligned cross domain data is expensive and may not even exist when task execution rates differ or there exists systematic embodiment mismatch between the domains. For example, a child can imitate an adult running, but not achieve the same speed. Our set up emulates a natural setting in which humans compare how they perform tasks to how other agents perform the same tasks in order to find structural similarities and identify domain correspondences. We now proceed to introduce a theoretical framework that explains how and when the DAIL problem can be solved by MDP alignment followed by adaptation.

%% file: sections/alignability.tex
\section{Alignable MDPs}
\label{section:alignability}
Let $\Pi^{*}_{\mc{M}}$ be the set of all optimal policies for MDP $\mc{M}$. We define an occupancy measure \citep{syed2008apprenticeship} $q_{\pi}: \S \times \A \rightarrow \sR$ for policy $\pi$ executed in MDP $\mc{M}$ as $q_{\pi}(s, a) = \pi(a | s) \sum_{t=0}^{\infty} \gamma^{t} \Pr(s^{(t)} = s; \pi, \mc{M})$. We further define the optimality function $O_{\mx}: \sx \times \ax \rightarrow \{0, 1\}$ for an MDP $\mx$ as $O_{\mx}(s_{x}, a_{x}) = 1$ if $\exists \pi_{x}^* \in \Pi^{*}_{\mx}$ such that $(s_{x}, a_{x}) \in \mathrm{supp}(q_{\pi_{x}^*})$ and $O_{\mx}(s_{x}, a_{x}) = 0$ otherwise. We are now ready to formalize MDP reductions: a class of structure preserving maps between MDPS. 

\begin{figure}[t]
  \begin{center}
    \includegraphics[width=0.45\textwidth]{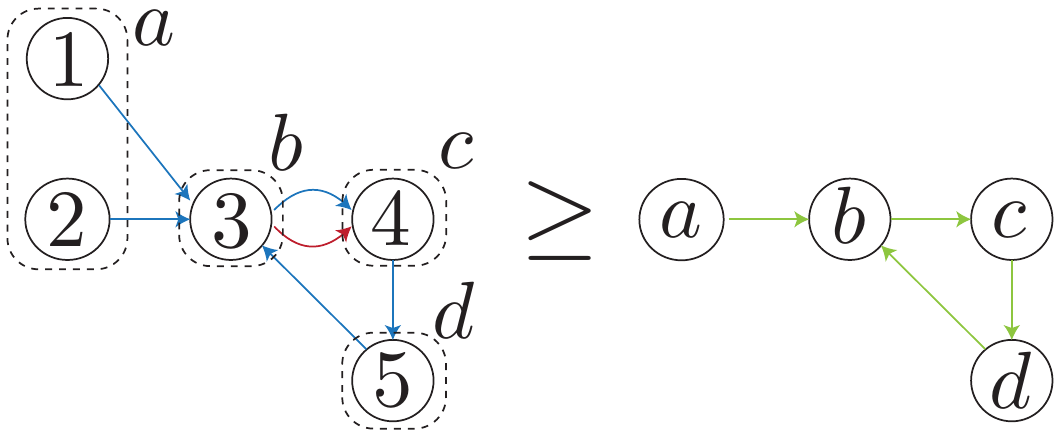}
  \end{center}
  \vspace{-10pt}
  \caption{\textbf{MDP Reduction Example} between execution traces in $\mx$ (Left) and $\my$ (Right), where $\mx \geq \my$. States correspond to nodes and actions to colors. The shown reduction merges nodes in dotted boxes to their corner label, e.g $\phi(1) = \phi(2) = a$, and both blue, red actions in $\mx$ to the green action in $\my$. }
  \vspace{-10pt}
  \label{fig:isomorphism} 
\end{figure}

\begin{restatable}{definition}{mdpreduction} An \textbf{MDP reduction} from $\mx = (\sx, \ax, \px, \eta_{x}, \rx)$ to $\my = (\sy, \ay, \py, \eta_{y}, \ry)$ is a tuple $r = (\phi, \psi)$ where $\phi: \sx \to \sy, \psi: \ax \to \ay$ are maps that preserve: 
\vspace{-5pt}
\begin{enumerate}
\thinmuskip=0mu
\item ($\pi$-optimality) $\forall (s_{x}, a_{x}, s_{y}, a_{y}) \in \sx \times \ax \times \sy \times \ay:$ 
\begin{flalign}
&O_{\my}(\phi(s_{x}),\psi(a_{x})) = 1\Rightarrow O_{\mx}(s_{x},a_{x})=1&\label{eq:reward}\\
&O_{\my}(s_{y}, a_{y}) = 1 \Rightarrow \phi^{\minus1}(s_{y}) \neq \emptyset, \psi^{\minus1}(a_{y}) \neq \emptyset \label{eq:surjective} 
\end{flalign}
\vspace{-15pt}
\item (dynamics) $\forall (s_{x}, a_{x}, s_{y}, a_{y}) \in \sx \times \ax \times \sy \times \ay$ such that $O_{\my}(s_{y}, a_{y}) = 1, s_{x} \in \phi^{-1}(s_{y}), a_{x} \in \psi^{-1}(a_{y}):$ 
\begin{align}
    \py(s_{y}, a_{y}) = \phi(\px(s_{x}, a_{x})) \label{eq:transition}
\end{align}
\end{enumerate}
where we define $\phi^{-1}(s_{y}) = \{ s_{x} | \phi(s_{x}) = s_{y} \}$,  $\psi^{-1}(a_{y}) = \{ a_{x} | \psi(a_{x}) = a_{y} \}$. 
Furthermore, $r$ is an \textbf{MDP permutation} if and only if $\phi, \psi$ are bijective.
\end{restatable}
In words, Eq. \ref{eq:reward} states that only optimal state, action pairs in $x$  can be mapped to optimal state, action pairs in $y$ and Eq.  \ref{eq:surjective} states that $r$ must be surjective on the set of optimal state, action pairs in $y$. These properties imply that a policy in $\mx$ should have more flexibility choosing optimal actions than one in $\my$. Eq. \ref{eq:transition} states that a reduction must preserve (deterministic) dynamics. We use the notation $\mx \geq_{\phi, \psi} \my$ to denote that $(\phi, \psi)$ is a reduction from $\mx$ to $\my$, and the shorthand $\mx \geq \my$ to denote that $\mx$ reduces to $\my$.
To gain an intuitive understanding of MDP reductions, picture the execution trace of an optimal policy as a directed graph with colored edges in which the nodes correspond to states visited by an optimal policy, and the colored edges correspond to actions taken. An MDP reduction from $\mx$ to $\my$ homomorphs the execution graph of an optimal policy in $\mx$ to a execution graph of an optimal policy in $\my$. Figure  \ref{fig:isomorphism} shows an example of a valid reduction from $\mx$ to $\my$: states $1, 2$ in $\sx$ are mapped (merged) to state $a$ in $\sy$ and the blue, red actions in $\ax$ are mapped to the green action in $\ay$. Intuitively, if $\mx \geq_{\phi, \psi} \my$, then $(\phi, \psi)$ compresses $\mx$ by merging all optimal state, action pairs that have identical dynamics properties.
\begin{restatable}{definition}{alignability} 
\label{def:alignability}
Two MDPs $\mx, \my$ are \textbf{alignable} if and only if $\mx \geq \my$ or $\my \geq \mx$.
\end{restatable}
\vspace{-5pt}
Definition \ref{def:alignability} states that MDPs are alignable if reductions exists between them, meaning that they share structure. We use  $\Gamma(\mx, \my) = \{(\phi, \psi) | \mx \geq_{\phi, \psi} \my\}$ to denote the set of all valid reductions from $\mx$ to $\my$. Reductions have a particularly useful property which is that they adapt policies across alignable MDPs. Consider a state map $f: \sx \rightarrow \sy$, an inverse action map $g: \ay \rightarrow \ax$, and a composite policy $\hat{\pi}_{x} = g \circ \pi_{y} \circ f$ (see Figure \ref{fig:setup}(b)). In words, $\hat{\pi}_{x}$ maps a self state to an expert state via $f$, simulates the expert’s action choice for the mapped state via $\pi_y$, then chooses a self action that corresponds to the simulated expert action with $g$. The following lemma holds for $\hat{\pi}_x$.

\begin{restatable}{theorem}{piadapt}
\label{lemma:pi_adapt}
Let $\mx, \my$ be MDPs satisfying Assumption \ref{ass:regularity} (see Appendix \ref{sec:proofs}), $\mx \geq_{\phi, \psi} \my$, and $\pi_{y}$ be optimal in $\my$. Then, $\forall g: \ay \rightarrow \ax$ s.t $\psi \circ g(a_y) = a_y ~~ \forall a_{y} \in \{a_{y} | \exists s_{y} \in \sy~s.t~O_{\my}(s_{y}, a_{y}) = 1\}$, it holds that $\hat{\pi}_{x} = g \circ \pi_{y} \circ \phi$ is optimal in $\mx$.
\end{restatable}
\vspace{-5pt}
Theorem \ref{lemma:pi_adapt} states that the state, action maps $(f, g^{-1})$ chosen to be a reduction can adapt optimal policies between alignable MDPs. Here onwards we interchangeably refer to $(f, g)$ as "alignments". We now show how the DAIL problem can be solved by first solving an MDP alignment problem followed by an adaptation step. 

\begin{figure}[t]
  \begin{center}
    \includegraphics[width=0.2\textwidth]{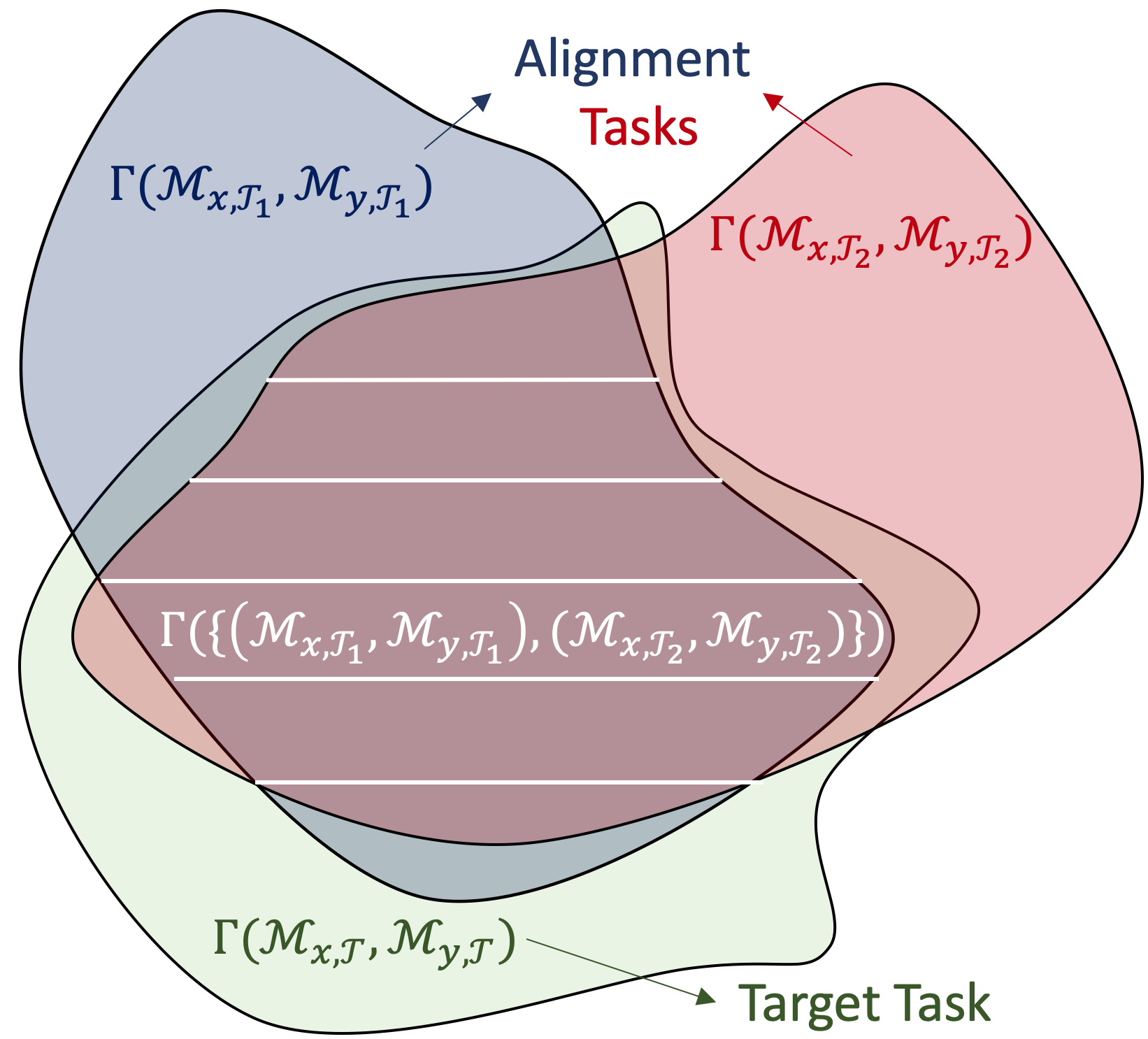}
  \end{center}
  \vspace{-10pt}
  \caption{\textbf{MDP Alignment Problem}. Blue/red, and green regions denote sets of MDP alignments for two alignment tasks and the target task, respectively. White hatches cover the solution set to the MDP alignment problem which is the intersection of all sets}
  \vspace{-10pt}
  \label{fig:region}
\end{figure}

\begin{restatable}{definition}{compatibility}
\label{def:jointalign}
Let $(\mx, \my), (\mx', \my') \in \Omega^2$ be two MDP pairs. Then, $(\mx, \my) \sim (\mx', \my')$, i.e they are \textbf{joint alignable}, if $\Gamma(\mx, \my) \cap \Gamma(\mx', \my') \neq \emptyset$.  
\end{restatable}
\vspace{-5pt}
In words, two MDP pairs are joint alignable if there exists a shared reduction. We define an equivalence class $[(\mx, \my)]_{\sim} = \{(\mx', \my')~~|~~(\mx', \my') \sim (\mx, \my)\}$ of MDP pairs that share reductions. Overloading notation, $\Gamma(\{(\mx^{i}, \my^{i})\}_{i=1}^{N}) = \{(\phi, \psi)~|~(\phi, \psi) \in \Gamma(\mx^{1}, \mx^{1}) \cap \cdots \cap \Gamma(\mx^{N}, \mx^{N})\}$. 
We now formally state the MDP alignment problem: Let $(\mx_{, \T}, \my_{, \T})$ be an MDP pair for a target task $\T$. Given an alignment task set $\mc{D}_{x, y} = \{(\mc{D}_{\mx_{, \T_{i}}}, \mc{D}_{\my_{, \T_{i}}})\}_{i = 1}^{N}$ comprising unpaired, unaligned demonstrations for MDP pairs $\{(\mx_{, \T_{i}}, \my_{, \T_{i}})\}_{i = 1}^{N} \subseteq [(\mx_{, \T}, \my_{, \T})]_{\sim}$, determine $(\phi, \psi) \in \Gamma(\{(\mx_{, \T_{i}}, \my_{, \T_{i}})\}_{i = 1}^{N})$ such that $(\phi, \psi) \in \Gamma(\mx_{, \T}, \my_{, \T})$. As shown in Figure \ref{fig:region}, with more MDP pairs, there are likely a smaller the number of joint alignments $|\Gamma(\{(\mx_{, \T_{i}}, \my_{, \T_{i}})\}_{i = 1}^{N})|$ and, as a result, $(\phi, \psi) \in \Gamma(\{(\mx_{, \T_{i}}, \my_{, \T_{i}})\}_{i = 1}^{N})$ is more likely to "generalize" to an MDP pair for a new target task $(\mx_{, \T}, \my_{, \T})$ in the equivalence class. Analogously, in a standard supervised learning problem, more training data is likely to shrink the set of models performing optimally on the training set but poorly on the test set. We can then use $(\phi, \psi)$ for DAIL: given cross domain demonstrations $\mc{D}_{\my_{, \T}}$ for the target task $\T$, learn an expert domain policy $\pi_{y, \T}$, and adapt it into the self domain using $(\phi, \psi)$ according to Theorem \ref{lemma:pi_adapt}. 

We can now assess when domains with embodiment and viewpoint mismatch have meaningful state correspondences, i.e MDP reductions, thus allowing for domain adaptive imitation. The states of a human expert with more degrees of freedom than a robot imitator can be merged into the robot states if the task only requires the robot's degrees of freedom and the execution traces share structure, e.g traces are both cycles. However, if the task requires all degrees of freedom possessed only by the human, the robot cannot find meaningful correspondences, and also cannot imitate the task. Two MDPs for different viewpoints of an agent performing a task are MDP permutations since there is a one-to-one correspondence between state, actions at same timestep in the execution trace of an optimal policy.

\section{Learning MDP Reductions}
\label{section:discovery}
We now derive objectives that can be optimized to learn MDP reductions. We propose distribution matching and policy performance maximization. We first define the distributions to be matched. 
\begin{definition}
\label{def:co-domain-exec}
Let $\gM_x$, ${\gM_y}$ be two MDPs and $\hat{\pi}_{x} = g \circ \pi_{y} \circ f$ for $f: \gS_x \to {\gS_y}, g: {\gA_y} \to {\gA_x}$ and policy $\pi_y$. The \textbf{co-domain policy execution process} $\gP_{\hat{\pi}_{x}} = \{\hat{s}_{y}^{(t)}, \hat{a}_{y}^{(t)}\}_{t \geq 0}$ is realized by running $\hat{\pi}_{x}$ in $\mx$, i.e:   
\begin{align}
    s^{(0)}_x \sim \eta_{x},~~\hat{s}_y^{(t)} = f(s_x^{(t)}),~~ \hat{a}_y^{(t)} \sim {\pi}_y(\cdot | \hat{s}_{y}^{(t)}), \notag\\
    a_x^{(t)} = g(\hat{a}_{y}^{(t)}),~~s_{x}^{(t+1)} = \px(s_x^{(t)}, a_x^{(t)}) \quad \forall t \geq 0 \label{eq:policy}
\end{align}
\end{definition}
\vspace{-5pt}
The target distribution $\sigyy$ is over transitions uniformly sampled from execution traces of $\pi_{y}$ in $\my$ and the proxy distribution $\sigxy$ is over cross domain transitions uniformly sampled from realizations of $\mc{P}_{\hat{\pi}_{x}}$, i.e running $\hat{\pi}_{x}$ in $\mx$.
{\thickmuskip=0mu
\begin{align}
&\sigyy(s_y, a_y, s'_y) :=\notag \\
&\lim_{T \to \infty} \frac{1}{T} \sum\limits_{t=0}^{T} \Pr(s^{(t)}_y = s_y, a^{(t)}_y = a_y, s^{(t+1)}_y = s'_y; \pi_{y})\\
&\sigxy(s_y, a_y, s'_y) :=\notag\\
&\lim_{T \to \infty} \frac{1}{T} \sum\limits_{t=0}^{T} \Pr(\hat{s}_y^{(t)} = {s}_y, \hat{a}_y^{(t)} = {a}_y, \hat{s}_y^{(t+1)} = {s}'_y; \gP_{\hat{\pi}_{x}})
\end{align}}
We now propose two concrete objectives to optimize for: 1. optimality of $\hat{\pi}_{x}$, 2. $\sigxy = \sigyy$. In other words, we seek to learn $f, g$ that matches distributions over transition tuples in domain $y$ while maximizing policy performance in domain $x$. The former captures the dynamics preservation property from Eq. \ref{eq:transition} and the latter captures the optimal policy preservation property from Eq. \ref{eq:reward}, \ref{eq:surjective}. The following theorem uncovers the connection between our objectives and MDP reductions.


\begin{restatable}{theorem}{main}
\label{thm:main}
Let $\mx, \my$ be MDPs satisfying Assumption \ref{ass:regularity} (see Supp Materials). If $\mx \geq \my$, then $\exists f: \gS_x \to \gS_y, g: \gA_y \to \gA_x$, and an optimal covering policy $\pi_y$ (see Appendix \ref{sec:proofs}) that satisfy objectives 1 and 2. Conversely, if $\exists f: \gS_x \to \gS_y$, an injective map $g: \gA_y \to \gA_x$ and an optimal covering policy $\pi_y$ satisfying objectives 1 and 2, then $\mx \geq \my$ and $\exists (\phi, \psi) \in \Gamma(\gM_x, \gM_y)$ s.t $f = \phi$ and $\psi \circ g(a_y) = a_y, \forall a_y \in \gA_y$.
\end{restatable}
\vspace{-5pt}
Theorem \ref{thm:main} states that if two MDP are alignable, then objectives 1 and 2 can be satisfied. Conversely, if 1 and 2 can be satisfied for two MDPs with state map $f$ and an injective action map $g$, then the MDPs must be alignable and all solutions $(f, g)$ are reductions. While Theorem \ref{thm:main} requires that MDPs be alignable to guarantee identifiability of solutions obtained via optimizing for objectives 1 and 2, our experiments will also run on MDPs that are "weakly" alignable, i.e. Eq. \ref{eq:reward}, \ref{eq:surjective}, \ref{eq:transition} do not hold exactly, but intuitively share structure. In the next section, we derive a simple algorithm to learn MDP reductions.

%% file: sections/algorithm.tex
\begin{algorithm*}[b]
\SetAlgoLined
\textbf{input}: Alignment task set $\mc{D}_{x, y} = \{(\mc{D}_{\mx_{, \T_{i}}}, \mc{D}_{\my_{, \T_{i}}})\}_{i = 1}^{N}$ of unpaired trajectories, fitted $\pi_{y, \T_{i}}^{*}$ \\
\textbf{while} not done \textbf{do}: \\
~~~\textbf{for} $i = 1, ..., N$ \textbf{do}: \\
~~~~~~ Sample $(s_{x}, a_{x}, s_{x}') \sim \mc{D}_{\mx_{, \T_{i}}}, (s_{y}, a_{y}, s_{y}') \sim \mc{D}_{\my_{, \T_{i}}}$ and store in buffer $\mc{B}_{x}^{i}, \mc{B}_{y}^{i}$\\
~~~~~~\textbf{for} $j = 1, ..., M$ \textbf{do}: \\
~~~~~~~~~ Sample mini-batch $j$ from $\mc{B}^{i}_{x}, \mc{B}^{i}_{y}$ \\
~~~~~~~~~ Update dynamics model with: $-\hat{\E}_{\pi^*_{x, \T_{i}}}[\nabla_{\theta_{P}}(P^{x}_{\theta_{P}}(s_{x}, a_{x}) - s_{x}')^2]$  \\
~~~~~~~~~ Update discriminator: $\hat{\E}_{\pi^*_{y, \T_{i}}}[\nabla_{\theta_{D}^{i}} \log D_{\theta_{D}^{i}}(s_{y}, a_{y}, s_{y}')] + \hat{\E}_{\pi^*_{x, \T_{i}}}[\nabla_{\theta_{D}^{i}} \log \big(1 - D_{\theta_{D}^{i}}(\hat{s}_{y}, \hat{a}_{y}, \hat{s}_{y}')\big)]$ \\
~~~~~~~~~ Update alignments $(f_{\theta_{f}}, g_{\theta_{g}})$ with gradients:
\vspace{-5pt}
\begin{align*}
-\hat{\E}_{\pi^*_{x, \T_{i}}}[\nabla_{\theta_{f}} \log D_{\theta_{D}}(\hat{s}_{y}, \hat{a}_{y}, \hat{s}_{y}')] + \hat{\E}_{\pi^*_{x, \T_{i}}}[\nabla_{\theta_{f}}(\hat{\pi}_{x, \T_{i}}(s_{x}) - a_{x})^2]\\
-\hat{\E}_{\pi^*_{x, \T_{i}}}[\nabla_{\theta_{g}} \log D_{\theta_{D}}(\hat{s}_{y}, \hat{a}_{y}, \hat{s}_{y}')] + \hat{\E}_{\pi^*_{x, \T_{i}}}[\nabla_{\theta_{g}}(\hat{\pi}_{x, \T_{i}}(s_{x}) - a_{x})^2]
\end{align*}
\vspace{-10pt}
\SetAlgoLined
\caption{Generative Adversarial MDP Alignment (GAMA)}
\label{algo: GAMA}
\end{algorithm*}

\begin{figure*}[t]
\centering
\includegraphics[width=1\textwidth]{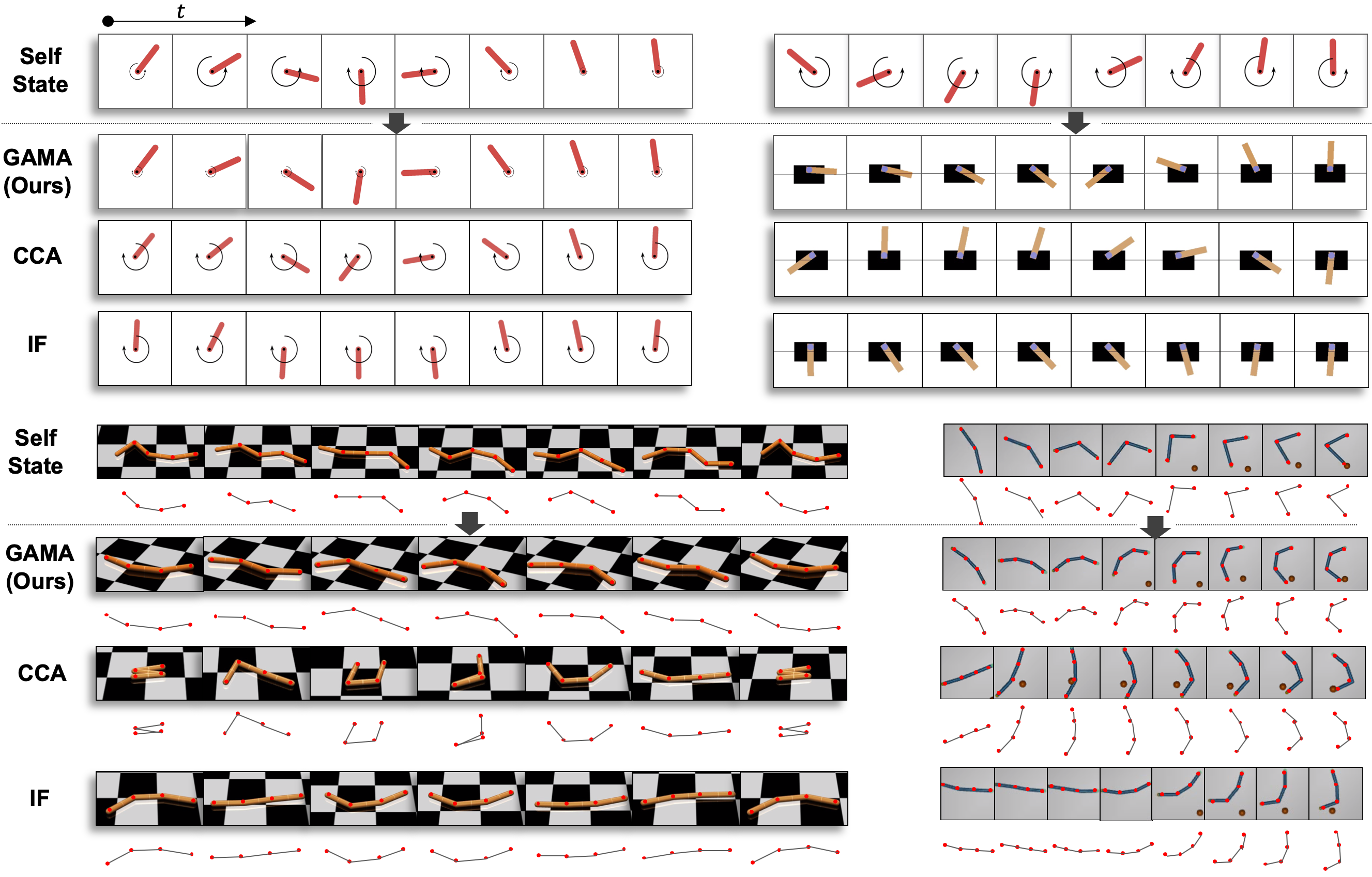}
\vspace{-25pt}
\caption{\textbf{MDP Alignment Visualization}. The state maps learned by GAMA and two representative baselines - CCA and IF - are shown for pen$\leftrightarrow$pen (Top Left), pen$\leftrightarrow$cart (Top Right), snake4$\leftrightarrow$snake3 (Bottom Left), reach2$\leftrightarrow$reach3 (Bottom Right). See Appendix \ref{sec:app:visualization} to see more baselines. GAMA is able to recover MDP permutations for alignable pairs pen$\leftrightarrow$pen, pen$\leftrightarrow$cart and find meanigful correspondences between "weakly alignable" pairs snake4$\leftrightarrow$snake3, reach2$\leftrightarrow$reach3. See \url{ https://youtu.be/l0tc1JCN_1M} for videos}
\vspace{-10pt}
\label{fig:mappings}
\end{figure*}

\section{Generative Adversarial MDP Alignment}
\label{section:algorithm}
Building on Theorem~\ref{thm:main}, we propose the following training objective for aligning MDPs: 
\begin{align}
    \min_{f, g} -J(\hat{\pi}_{x}) + \lambda d(\sigxy, \sigyy)
\end{align}
where $J(\hat{\pi}_{x})$ is the performance of $\hat{\pi}_{x}$, $d$ is a distance metric between distributions, and $\lambda > 0$ is a Lagrange multiplier. In practice, we found that injectivity of $g$ is unnecessary to enforce in continuous domains. We now present an instantiation of this framework: Generative Adversarial MDP Alignment (GAMA). 
Recall that we are given an alignment task set $\mc{D}_{x, y} = \{(\mc{D}_{\mx_{, \T_{i}}}, \mc{D}_{\my_{, \T_{i}}})\}_{i = 1}^{N}$. In the alignment step, we learn $\pi^*_{y, \T_{i}}, \forall \T_{i}$ and parameterized state, action maps $f_{\theta_{f}}: \sx \rightarrow \sy, g_{\theta_{g}}: \ay \rightarrow \ax$ that compose $\hat{\pi}_{x, \T_{i}} = g_{\theta_{g}} \circ \pi^*_{y, \T_{i}} \circ f_{\theta_{f}}$. To match $\sigxy, \sigyy$, we employ adversarial training \citep{goodfellow2014generative} in which separate discriminators $D_{\theta_{D}^{i}}$ per task are trained to distinguish between "real" transitions $(s_{y}, a_{y}, s_{y}') \sim \pi_{y, \T_{i}}^{*}$ and "fake" transitions $(\hat{s}_{y}, \hat{a}_{y}, \hat{s}_{y}') \sim \hat{\pi}_{x, \T_{i}}$, where $\hat{s}_{y} = f_{\theta_{f}}(s_{x}), \hat{a}_{y} = \pi_{y}(\hat{s}_{y}), \hat{s}_{y}' = f_{\theta_{f}}(P^{x}_{\theta_{P}}(s_{x}, g(\hat{a}_{y})))$, and $P^{x}_{\theta_{P}}$ is a fitted model of the $x$ domain dynamics. (see Figure \ref{fig:setup}(b)) The generator, consisting of $f_{\theta_{f}}, g_{\theta_{g}} $, is trained to fool the discriminator while maximizing policy performance. The distribution matching gradients are back propagated through the learned dynamics, $\pi^*_{y, \T_{i}}$ is learned by Imitation Learning (IL) on $\mc{D}_{\my_{, \T_{i}}}$, and the policy performance objective on $\hat{\pi}_{x, \T_{i}}$ is achieved by IL on $\mc{D}_{\mx_{, \T_{i}}}$. In this work, we use behavioral cloning \citep{pomerleau1991efficient} for IL. We aim to find a saddle point $\{f, g\} \cup \{D_{\theta_{D}^{i}}\}_{i = 1}^{N}$ of the following objective: 
\vspace{-10pt}
\begin{align}
    \min_{f, g} \max_{\{D_{\theta_{D}^{i}}\}} \sum\limits_{i = 1}^{N} &\big(\E_{s_{x} \sim \pi^*_{x, \T_{i}}}[D_{KL}(\pi^*_{x, \T_{i}}(\cdot | s_{x}) || \hat{\pi}_{x, \T_{i}}(\cdot | s_{x}))] \nonumber \\[-7pt]
    &+ \lambda (\bb{E}_{\pi^*_{y, \T_{i}}}[\log D_{\theta_{D}^{i}}(s_y, a_y, s'_y)] \nonumber \\
    &+ \bb{E}_{\pi^*_{x, \T_{i}}}[\log (1 - D_{\theta_{D}^{i}}(\hat{s}_y, \hat{a}_y, \hat{s}_y'))] \big)
\end{align} 
where $D_{KL}$ is the KL-divergence. We provide the execution flow of GAMA in Algorithm \ref{algo: GAMA}. In the adaptation step, we are given expert demonstrations $\mc{D}_{\my_{, \T}}$ of a new target task $\T$, from which we fit an expert domain policy $\pi_{y, \T}$ which are composed with the learned alignments to construct an adapted self policy $\hat{\pi}_{x, \T} = g_{\theta_{g}} \circ \pi_{y, \T} \circ f_{\theta_{f}}$. We also experiment with a demonstration adaptation method which additionally trains an inverse state map $f^{-1}: \sy \rightarrow \sx$, adapts demonstrations $\mc{D}_{\my_{, \T}}$ into the self domain via $f^{-1}, g$, and applies behavioral cloning on the adapted demonstrations. (see Figure \ref{fig:setup}(c)) Notably, our entire procedure does not require paired, aligned demonstrations nor an RL step.

%% file: sections/relatedworks.tex
\textbf{Related Works}: 
Closely related to DAIL, the field of cross domain transfer learning in the context of RL has explored approaches to use state maps to exploit cross domain demonstrations in a pretraining procedure for a new target task for which self domain reward function is available. Canonical Correlation Analysis (CCA) \citep{Hotelling1936CCA} finds invertible projections into a basis in which data from different domains are maximally correlated. These projections can then be composed to obtain a direct correspondence map between states. \cite{Ammar2015Unsupervised, Joshi2018Target} have utilized an unsupervised manifold alignment (UMA) algorithm which finds a linear map between states with similar local geometric properties. UMA assumes the existence of hand crafted features along with a distance metric between them. This family of work commonly uses a linear statemap to define a time-step wise transfer reward and executes an RL step on the new task. Similar to our work, these works use an alignment task set of unpaired, unaligned trajectories to compute the state map. Unlike these works, we learn maps that preserve MDP structure, use deep neural network state, action maps, and achieve zero-shot transfer to the new task without an RL step.
More recent work in transfer learning across embodiment \citep{Gupta2017Learning} and viewpoint \citep{liu2018fromobs, sermanet2018tcn} mismatch obtain state correspondences from an alignment task set comprising paired, time-aligned demonstrations and use them to learn a state map or a state encoder to a domain invariant feature space. In contrast to this family of prior work, our approach learns both state, action maps from \emph{unpaired, unaligned} demonstrations. Also, we remove the need for additional environment interactions and an expensive RL procedure on the target task by leveraging the action map for zero-shot imitation. \cite{stadie2017third} have shown promise in using domain confusion loss and generative adversarial imitation learning \citep{ho2016generative} for learning across small viewpoint mismatch without an alignment task set, but fails in dealing with large viewpoint differences. Unlike \cite{stadie2017third}, we leverage the alignment task set to succeed in imitating across larger viewpoint mismatch and do not require an RL procedure. Some recent works \citep{liu2020statealignment} have proposed matching only state occupancy measures for imitation across dynamics mismatch. We compare our method to an appropriate distillation of such methods. MDP homomorphisms \citep{Ranvindran2002Homo} have been explored with the aim of compressing state, action spaces to facilitate planning. In similar vein, related works have proposed MDP similarity metrics based on bisimulation methods \citep{Ferns2004Bisumulation} and Boltzmann machine reconstruction error \citep{Ammar2014Automated}. While conceptually related to our MDP alignability theory, these works have not proposed scalable procedures to discover the homomorphisms and have not drawn connections to domain adaptive learning.

%% file: sections/experiments.tex
\begin{figure*}[t]
    \centering
    \includegraphics[width=1\textwidth]{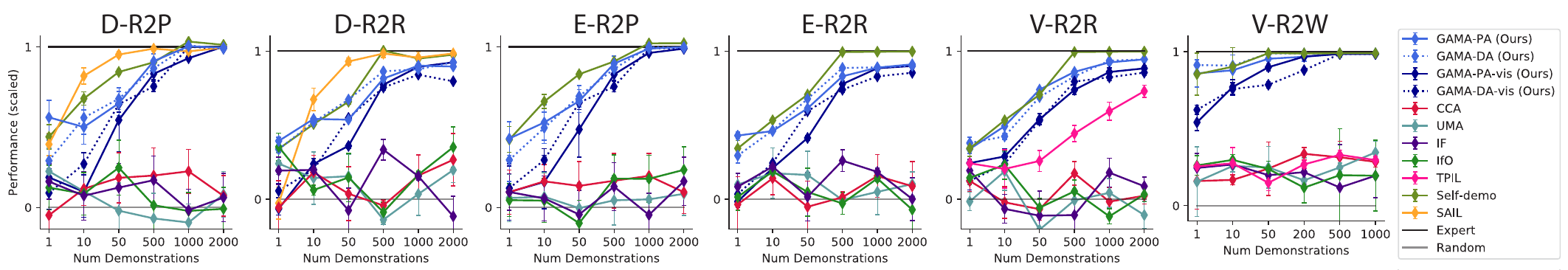}
    \vspace{-20pt}
    \caption{\textbf{Adaptation Complexity}. Notably, adaptation complexity of GAMA is close to that of the Self-demo baseline. Baselines fail at DAIL, mostly due to failing the alignment step. Results are averaged across 5 runs.}
    \label{fig:adaptation_complexity}
    \vspace{-15pt}
\end{figure*}

\section{Experiments}
\label{section:experiments}
Our experiments were designed to answer the following questions: (1). Can GAMA uncover MDP reductions? (2). Can the learned alignments $(f_{\theta_{f}}, g_{\theta_{g}})$ be leveraged to succeed at DAIL? We propose two metrics to measure DAIL performance. First, \emph{alignment complexity} which is the number of MDP pairs, i.e number of tasks, in the alignment task set needed to learn alignments that enable zero-shot imitation, given ample cross domain demonstrations for the target tasks. Second, \emph{adaptation complexity} which is the amount of cross domain demonstrations for the target tasks needed to successfully imitate tasks in the self domain without querying the target task reward function, given a sufficiently large alignment task set. Note that we include experiments with MDP pairs that are \emph{not perfectly alignable}, yet intuitively share structure, to show general applicability of GAMA for DAIL. We experiment with environments which are extensions of OpenAI Gym \citep{brockman2017openaigym}: pen, cart, reacher2, reacher3, reach2-tp, snake3, and snake4 denotes the pendulum, cartpole, 2-link reacher, 3-link reacher, third person 2-link reacher, 3-link snake, and 4-link snake environments, respectively. (self domain) $\leftrightarrow$ (expert domain) specify an MDP pair in the alignment task set. Model architectures and environment details are further described Appendix \ref{supp: arch}, \ref{supp: baseline}, \ref{supp: env}. We study two ablations of GAMA and compare against the following baselines: 


\textbf{GAMA - Policy Adapt (GAMA-PA)}: learns alignments by Algorithm \ref{algo: GAMA}, fits an expert policy $\pi_{y, \T}$ to $\mc{D}_{\my_{, \T}}$ for a new target task $\T$ and zero-shot adapts $\pi_{y, \T}$ to the self domain via $\hat{\pi}_{x, \T} = g_{\theta_{g}} \circ \pi_{y, \T} \circ f_{\theta_{f}}$. 

\textbf{GAMA - Demonstration Adapt (GAMA-DA)}: trains $f^{-1}$ in addition to Algorithm \ref{algo: GAMA}, adapts $\mc{D}_{\my_{, \T}}$ into the self domain via $(f^{-1}, g)$, and fits a self domain policy on the adapted demonstrations. 

\textbf{Self Demonstrations (Self-Demo)}: We behavioral clone on self domain demonstrations of the target task. This baseline sets an "upper bound" for the adaptation complexity. 

\textbf{Canonical Correlation Analysis (CCA) \citep{Hotelling1936CCA}}: finds invertible linear transformations to a space where domain data are maximally correlated when given unpaired, unaligned demonstrations.

\textbf{Unsupervised Manifold Alignment (UMA) \citep{Ammar2015Unsupervised}}: finds a map between states that have similar local geometries from unpaired, unaligned demonstrations. 

\textbf{Invariant Features (IF) \citep{Gupta2017Learning}}: finds invertible maps onto a feature space given state pairings. Dynamic Time Warping \citep{muller2007dtw} is used to obtain the pairings.

\textbf{Imitation from Observation (IfO) \citep{liu2018fromobs}}: learns a statemap conditioned on a cross domain observation given state pairings. Dynamic Time Warping \citep{muller2007dtw} is used to obtain the pairings. 

\textbf{Third Person Imitation Learning (TPIL) \citep{stadie2017third}}: simultaneously learns a domain agnostic feature space while matching distributions in the feature space. 

\textbf{State-Alignment Imitation Learning \citep{liu2020statealignment}}: Distribution matching imitation learning with a state occupancy matching objective. 

\begin{figure*}[t]
    \centering
    \includegraphics[width=1\textwidth]{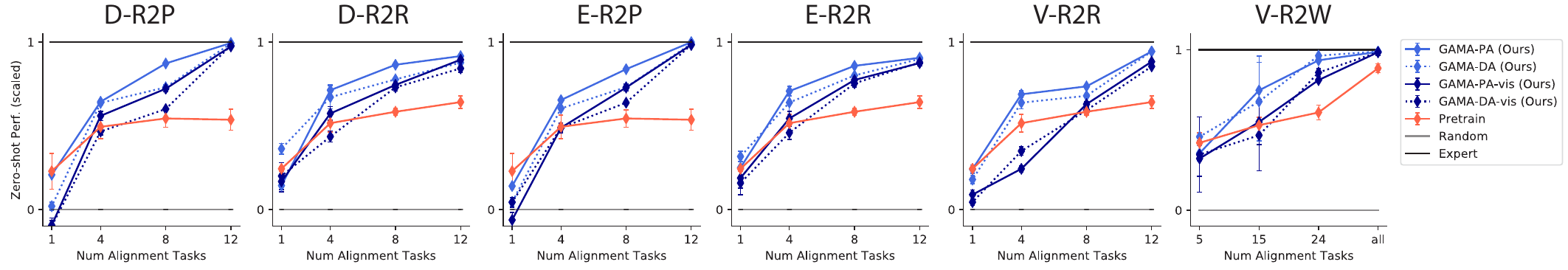}
    \vspace{-20pt}
    \caption{\textbf{Alignment Complexity}. Baselines cannot perform zero-shot imitation. Pretrain baseline shows the zero-shot performance of a policy directly pretrained on the self domain alignment tasks, when possible. Results are averaged across 5 runs.}
    \label{fig:alignment_complexity}
    \vspace{-10pt}
\end{figure*}

\subsection{MDP Alignment Evaluation}
Figure \ref{fig:mappings} visualizes the learned state map $f_{\theta_{f}}$ for several MDP pairs. The pen $\leftrightarrow$ pen alignment task (Figure \ref{fig:mappings}, Top Left) and reach$\leftrightarrow$reach-tp (Table \ref{table:identity}) task exemplify MDP pairs that are permutations. Similarly, the pen $\leftrightarrow$ cart alignment task (Figure \ref{fig:mappings}, Top Right) has a reduction that maps the pendulum's angle and angular velocity to those of the pole, as the cart's position and velocity are redundant state dimensions once an optimal policy has been learned. Table \ref{table:identity} presents quantitative evaluations of these simple alignment maps. For pen$\leftrightarrow$pen and reach2$\leftrightarrow$reach2-tp we record the average $\ell_{2}$ loss between the learned statemap prediction and the ground truth permutation. As for pen$\leftrightarrow$cart, we do the same on the dimensions for the pole's angle and angular velocity. Both Figure \ref{fig:mappings} and Table \ref{table:identity} shows that GAMA is able to learn simple reductions while baselines mostly fail to do so. The key reason behind this performance gap is that many baselines \citep{Gupta2017Learning, liu2018fromobs} obtain state maps from time-aligned demonstration data using Dynamic Time Warping (DTW). However, the considered alignment tasks contains unaligned demonstrations with diverse starting states, up to 2x differences in demonstration lengths, and varying task execution rates. We see that GAMA also outperforms baselines that learn from unaligned demonstrations \citep{Hotelling1936CCA, Ammar2015Unsupervised} by learning maps that preserve MDP structure with more flexible neural network function approximators. For snake4 $\leftrightarrow$ snake3 and reach2 $\leftrightarrow$ reach3, the MDPs may not be perfectly alignable, yet they intuitively share structure. From Figure \ref{fig:mappings} (Bottom Left) we see that GAMA identically matches two adjacent joint angles of snake4 to the two joint angles of snake3 and the periodicity of the snake's wiggle is preserved. On reacher2$\leftrightarrow$reacher3, GAMA learns a state map that matches the first joint angles and states that have similar extents of contraction.

\subsection{DAIL Performance}
We evaluate DAIL performance on six problems that span embodiment, viewpoint, and dynamics mismatch scenarios. See Appendix \ref{supp: env} for further details on each problem.  

\textbf{Dynamics-Reach2Reach (D-R2R)}: Self domain is reach2 and expert domain is reach2 with isotropic gaussian noise injected into the dynamics. We use the robot's joint level state-action space. The $N$ alignment tasks are reaching for $N$ goals and the target tasks are reaching for $12$ new goals, placed maximally far away from the alignment task goals. 

\textbf{Dynamics-Reach2Push (D-R2P)}: Same as D-R2R except the target task is pushing a block to a goal location. 

\textbf{Embodiment-Reach2Reach (E-R2R)}: Self domain is reach2 and expert is reach3. Rest is the same as D-R2R.

\textbf{Embodiment-Reach2Push (E-R2P)}: Self domain is reach2 and expert is reach3. Rest is the same as D-R2P.

\textbf{Viewpoint-Reach2Reach (V-R2R)}: Self domain is reach2 and expert domain is reach2-tp1 that has the same "third person" view state space as that in \cite{stadie2017third} with a $30^{\circ}$ planar offset. We use the robot's joint level state-action space. The alignment/target tasks are the same as D-R2R. 

\textbf{Viewpoint-Reach2Write (V-R2W)}: Self domain is reach2 and expert domain is reach2-tp2 that has a different "third person" view state space with a $180^{\circ}$ axial offset. We use the robot's joint level state-action space. The $N$ alignment tasks are reaching for $N$ goals and the target task is tracing letters as fast as possible. The transfer task differs from the alignment tasks in two key aspects: the end effector must draw a straight line from a letter's vertex to vertex and not slow down at the vertices in order to trace.  

Alignment complexity on the six problems is shown in Figure \ref{fig:alignment_complexity}. GAMA (light blue) is able to learn alignments that enable zero-shot imitation on the target task, showing clear gains over a simple pretraining procedure (orange) on the self domain MDPs in the alignment task set. Other baselines require an additional expensive RL step and thus cannot zero-shot imitate. Figure \ref{fig:adaptation_complexity} shows the adaptation complexity. Notably, GAMA (light blue) produces adapted demonstrations of similar usefulness as self demonstrations (olive green). Most baselines fail to learn alignments from unpaired, unaligned demonstrations and as a result fails at DAIL. TPIL succeeds at V-R2R, but fails at V-R2W which has a significantly larger viewpoint mismatch than V-R2R. SAIL outperforms GAMA and even the self-Demo baseline, but it's important to note that SAIL uses the self domain environment simulator unlike GAMA and Self-Demo. 

\subsection{DAIL with Visual Inputs}
The non-visual environment experiments in the previous section demonstrate the limitations of the time-alignment assumptions made in prior work without confounding variables such as the difficulty of optimization in high-dimensional spaces. In this section, we introduce two more variants of our method, GAMA-PA-vis and GAMA-DA-vis, which demonstrate that \emph{GAMA scales to higher dimensional, visual environments} with $64\times64\times3$ image states. Specifically, we train a deep spatial autoencoder on the alignment task set to learn an encoder with the architecture from \cite{levine2016end}, then apply GAMA on the (learned) latent space. Results are shown in Figure \ref{fig:alignment_complexity}, \ref{fig:adaptation_complexity}. We see that the alignment and adaptation complexity of GAMA-PA-vis (dark-blue, solid), GAMA-DA-vis (dark-blue, dotted) are both similar to that of GAMA-DA (light blue, solid), GAMA-PA (light blue, dotted) and better than baselines trained with the robot's joint-level representation.

%% file: sections/conclusion.tex
\section{Discussion and future work}
\label{sec:futurework}
We've formalized Domain Adaptive Imitation Learning which encompasses prior work in transfer learning across embodiment \citep{Gupta2017Learning} and viewpoint differences \citep{stadie2017third, liu2018fromobs} along with a practical algorithm that can be applied to both scenarios. We now point out directions future work. Our MDP alignability theory is a first step towards formalizing possible shared structures that enable cross domain imitation. While we've shown that GAMA empirically works well even when MDPs are not perfectly alignable, upcoming works may explore relaxing the conditions for MDP alignability to develop a theory that covers an even wider range of real world MDPs. Future works may also try applying GAMA in the imitation from observations scenario, i.e actions are not available, by aligning observations with GAMA and applying methods from \cite{sermanet2018tcn, liu2018fromobs}. Finally, we hope to see future works develop principled ways design a minimal alignment task set, which is analogous to designing a minimal training set for supervised learning. 

%% file: sections/acknowledgements.tex
\section*{Acknowledgements}
This research was supported by Sony, NSF (\#1651565, \#1522054, \#1733686), ONR (N00014-19-1-2145), AFOSR (FA9550-19-1-0024).

%% file: sections/supplementary.tex
\section{High-level Comparison to Baselines}
\label{supp:highlevel}
\captionof{table}{Comparison of baselines by attributes demonstrated in the paper. The "No Act" column denotes whether or not the demonstrations need to contain actions.}
{\tiny
\begin{center}
\begin{sc}
 \begin{tabular}{ccccccc}
 \toprule
 & Unpaired Data & Zeroshot Imit. & Embod. Mismatch & Viewpoint Mismatch & Single-domain Demo. & No Act. \\
 \midrule
 TPIL \citep{stadie2017third} & \greencheck & \redx & \redx & \greencheck & \greencheck & \redx \\
 IF \citep{Gupta2017Learning} & \redx & \redx & \greencheck & \redx & \redx & \redx \\
 IfO \citep{liu2018fromobs} & \redx & \redx & \redx & \greencheck & \redx  & \greencheck \\
 TCN \citep{sermanet2018tcn} & \redx & \redx & \redx & \greencheck & \greencheck & \greencheck \\
 \textbf{GAMA (ours)} & \greencheck & \greencheck & \greencheck & \greencheck & \redx & \redx \\  
 \bottomrule
\end{tabular}
\end{sc}
\end{center}
}
We note that methods such as IF has potential to be applied to the viewpoint mismatch problem and IfO, TCN have the potential to be applied to the embodiment mismatch problem, albeit they were not shown in the paper. TCN has shown mappings between humans and robots can be learned. However they haven't shown that robots can use these mappings to learn from human demonstrations. Below we summarize the key differences between GAMA and the main baselines. 
\begin{enumerate}[wide, labelwidth=!, labelindent=0pt]
\item We propose an \emph{unsupervised MDP alignment} algorithm (GAMA) capable of learning state correspondences from \emph{unpaired, unaligned demonstrations} while \cite{Gupta2017Learning, liu2018fromobs, sermanet2018tcn} obtain these correspondences from paired, time-aligned trajectories. Our demonstrations have varying length (up to 2x difference) and diverse starting positions. Since good observation correspondences are prerequisites to the success of \cite{Gupta2017Learning, liu2018fromobs, sermanet2018tcn}, our work provide the missing ingredient. Future work could try learning alignments with GAMA, then apply methods from \cite{Gupta2017Learning, liu2018fromobs, sermanet2018tcn} to perform CDIL when action information is unavailable from demonstrations. 
\vspace{-3pt}
\item \emph{We remove the need for an expensive RL procedure on a new target task}, by leveraging action information for zero-shot imitation. By learning a composite self policy with both state and action maps, we obtain a near-optimal self policy on new tasks without any environment interactions while prior approaches \citep{Gupta2017Learning, liu2018fromobs, sermanet2018tcn} require an additional RL step that involves self domain environment interactions.  
\vspace{-3pt}
\item \emph{We use a single algorithm to address both the viewpoint and embodiment mismatch} which have previously been dealt with different solutions.
\end{enumerate}

\section{GAMA model architecture}
\label{supp: arch}
The state, action map $f_{\theta_{f}}, g_{\theta_{g}}$, inverse state map $f^{-1}_{\theta_{f^{-1}}}$, transition function $P^{x}_{\theta_{P}}$, and discriminators $\{D_{\theta^{i}_{D}}\}_{i = 1}^{N}$ are neural networks with hidden layers of size $(200, 200)$. The fitted policies $\{\pi_{y, \T_{i}}\}_{i = 1}^{N}$ for GAMA-PA and $\pi_{x, \T}$ for GAMA-DA all have hidden layers of size $(300, 200)$. All models are trained with Adam optimizers \citep{kingma2014adam} using decay rates $\beta_{1} = 0.9, \beta_{2} = 0.999$. For the spatial autoencoders used in GAMA-PA-img and GAMA-DA-img, we use the same architecture as in \cite{finn2015spatial} We use a learning rate of $1e$-$4$ for the alignment maps and $1e$-$5$ for all other components. These parameters are fixed across all experiments. 

\section{Baseline Implementation Details}
\label{supp: baseline}

In this section we describe our implementation details of the baselines. 

\textbf{Obtaining State Correspondences} We use 5000 sampled trajectories in both expert and self domains to learn the state map for IF and CCA. For UMA, we use 20 sampled trajectories to learn that in pendulum and cartpole environment and 50 trajectories in reacher, reacher-tp environment (much beyond these numbers UMA is computationally intractable). For IF and IfO, we use Dynamic Time Warping (DTW) \citep{muller2007dtw} to obtain state correspondences. For IF, DTW uses the (learned) feature space as a metric space to estimate the domain correspondences. For IfO, DTW is applied on the state space. We follow the implementation procedure in \cite{Gupta2017Learning}. 

To visualize and quantitatively evaluate the statemaps learned in prior work, we compose the encoder and decoder for IF and use the Moore-Penrose pseudo inverse of the embedding matrix for UMA and CCA. 

\textbf{Transfer Learning} In the transfer learning phase for CCA, UMA, IF, and IfO they define a proxy reward function on the target task by using the state correspondence map. 

\begin{equation*}
    r_{\text{proxy}}(s_x^{(t)}) = \frac{1}{|\Tau|} \sum_{\tau \in \Tau} {\|f(s_{y, \tau}^{(t)})-g(s_x^{(t)})\|_2^2}
\end{equation*}, where $s_x^{(t)}$ is a self domain state at time $t$, $\Tau$ is the collection of expert demonstrations, and $s_{y, \tau}^{(t)}$ is the expert domain state at time $t$ in trajectory $\tau$. IfO additionally defines a penalty reward for deviating from states encountered during training. We refer readers to their paper \citep{liu2018fromobs} for further details.
The transferability results of Figure \ref{fig:evaluation} (Right) show the learning curve for training on the ground truth reward for the target task where the policy is pretrained with a training procedure on the proxy reward. All RL steps are performed using the Deep Deterministic Policy Gradient (DDPG) \citep{lillicrap2015continuous} algorithm.

\textbf{Architecture} For UMA and CCA, the embedding dimension is the minimum state dimension between the expert and self domains. For UMA, we use one state sample every 5 timesteps to reduce the computational time, and we match the pairwise distance matrix of 3-nearest neighbors. 

For IF, we use 2 hidden layer with 64 hidden units each and leaky ReLU non-linearities to parameterize embedding function and decoders, the dimension of common feature space is set to be 64. The optimizer are same with respect to our models and the learning rate is $1e$-$3$. For IfO, we use the same architecture as the statemap in GAMA for their observation conditioned statemap. For TPIL, we use 3 hidden layer with 128 hidden units each and ReLU non-linearities to parameterize the feature extractor, classifier and domain classifier. We use Adam Optimizer with default decay rates and learning rate $1e$-$3$ to train the discriminator and use same optimizer and learning rate with respect to our model to train the policy.

\section{Environments and DAIL tasks}
\label{supp: env}
We use the 'Pendulum-v0', 'Cartpole-v0' environments for the pendulum and cartpole tasks which have state space $(w, \dot{w})$ and $(w, \dot{w}, x, \dot{x})$, respectively, where $w$ is the angle of the pendulum/pole and $x$ is the position of the cart. The action spaces are $(F_{w})$ and $(F_{x})$ where $F_{w}$ is the torque applied to the pendulum's pivot and $F_{x}$ is the x-direction force applied to the cart. For snake3, snake4 we use an extension \cite{wang2018nervenet} of the 'Swimmer-v0' environment from Gym \cite{brockman2017openaigym}. A $K$ link snake has a state representation $(w_{1}, ..., w_{K}, \dot{w_{1}}, ..., \dot{w_{K}})$ where $w_{k}$ is the angle of the $k^{th}$ snake joint. The action vector has the form $(F_{w_{1}}, ... F_{w_{K}})$ where $F_{w_{k}}$ is the torque applied to joint $k$. All reacher environments were extended from the 'Reacher-v0' gym environment. A $k$ link reacher has a state vector of the form $(c_{1}, ..., c_{K}, \dot{w_{1}}, ..., \dot{w_{K}}, x_{g}, y_{g})$ where $c_{k}$ is the coordinates of the $k^{th}$ reacher joint and $(x_{g}, y_{g})$ is the position of the goal. Note the key difference with the original Reacher-v0 environment is that we use coordinates of joints instead of joint angles and the difference vector between the end effector and the goal coordinate was removed from the state to make the task more challenging. The action vector has the form $(F_{w_{1}}, ... F_{w_{K}})$ where $F_{w_{k}}$ is the torque applied to joint $k$. Below we specifically describe each DAIL task. The statemap acts only on the non-goal dimensions. 

\textbf{Dynamics-Reach2Reach (D-R2R)}: Self domain is reach2 and expert domain is reach2 with isotropic gaussian noise injected into the dynamics. State, action spaces are the same the one for a $k$-link reacher with $k = 2$. The $N$ alignment tasks are reaching for $N$ goals near the wall of the arena and the target tasks are reaching for $12$ new goals near the corner of the arena. The new goals are placed as far as possible from the alignment task goals within the bounds of the arena to make the task more challenging.

\textbf{Dynamics-Reach2Push (D-R2P)}: Same as D-R2R except the target task is pushing a block to a goal location. State, action spaces are the same the one for a $k$-link reacher with $k = 2$. Here the goal location represents the location to push the block to. Block is always initialized in the same location. 

\textbf{Embodiment-Reach2Reach (E-R2R)}: Self domain is reach2 and expert is reach3. Rest is the same as D-R2R.

\textbf{Embodiment-Reach2Push (E-R2P)}: Self domain is reach2 and expert is reach3. Rest is the same as D-R2P.

\textbf{Viewpoint-Reach2Reach (V-R2R)}: Self domain is reach2 and expert domain is reach2-tp1 that has the same "third person" view state space as that in \cite{stadie2017third} with a $30^{\circ}$ planar offset. The state space is a projection of the joint coordinates onto the offset viewing plane, e.g a joint coordinate $(1, 1)$ in the self domain is corresponds to $(1, 0.7)$ in the expert domain. The alignment/target tasks are the same as D-R2R. 

\textbf{Viewpoint-Reach2Write (V-R2W)}: Self domain is reach2 and expert domain is reach2-tp2 that has a different "third person" view state space with a $180^{\circ}$ axial offset. Thus a joint coordinate $(1, 1)$ in the self domain corresponds to $(-1, -1)$ in the expert domain. We use the robot's joint level state-action space. The $N$ alignment tasks are reaching for $N$ goals and the target task is tracing letters as fast as possible. The goal location in the writing task represents the next vertex of the letter to trace. Once the first vertex is reached, the goal coordinates are updated to be the next vertex coordinates. The reward function is defind as follows: 
\vspace{-5pt}
\begin{align*}
    R_{write}(s) = 
    \begin{cases}
    100 \quad \text{if state $s$ corresponds to reaching a vertex} \\
    -1 \quad \text{else}
    \end{cases}
\end{align*}
Thus the agent must perform a sequential reaching task and accomplish it as fast as possible. The key difference with a normal reaching task is that the reacher must not slow down at each vertex and plan it's path accordingly in order to minimize drastic direction changes. Further more the reward is significantly more sparse than the original reacher reward which gets reward inversely proportional to the distance between the end effector and the goal. 

\section{MDP Alignment Visualization} 
\label{sec:app:visualization}
\begin{figure*}[h!]
\centering
\includegraphics[width=1\textwidth]{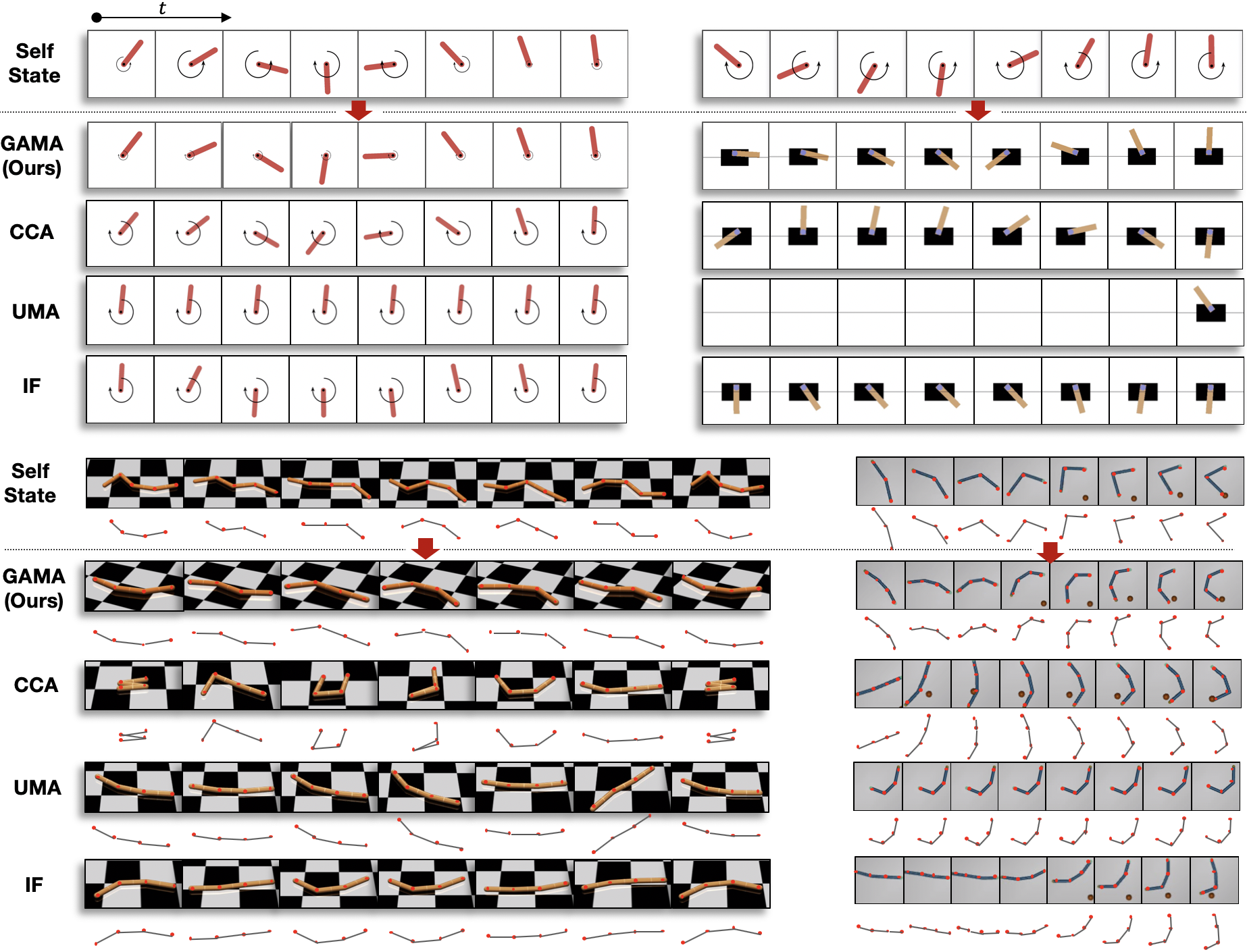}
\vspace{-25pt}
\caption{\textbf{MDP Alignment Visualization (Extended)}. The state maps learned by GAMA and baselines are shown for pen$\leftrightarrow$pen (Top Left), pen$\leftrightarrow$cart (Top Right), snake4$\leftrightarrow$snake3 (Bottom Left), reach2$\leftrightarrow$reach3 (Bottom Right). See Appendix \ref{sec:app:visualization} to see more baselines. GAMA is able to recover MDP permutations for alignable pairs pen$\leftrightarrow$pen, pen$\leftrightarrow$cart and find meaningful correspondences between "weakly alignable" pairs snake4$\leftrightarrow$snake3, reach2$\leftrightarrow$reach3. For pen$\leftrightarrow$cart, UMA learns a statemap that outputs out-of-bounds coordinates mainly because the pendulum demonstrations are concentrated around the pole upright state. The optimal UMA embedding matrix in this case is a zero matrix. Then the UMA state map matrix norm is proportional to the inverse embedding matrix norm which is very large. See \url{ https://youtu.be/l0tc1JCN_1M} for videos}
\vspace{-10pt}
\label{fig:mappings_supp}
\end{figure*}


\newpage

\input{sections/proofs}

%% file: sections/proofs.tex
\section{Proofs}
\label{sec:proofs}


We start by introducing definitions and assumptions that will be used in proving both Theorem \ref{lemma:pi_adapt}, \ref{thm:main}

\begin{restatable}{definition}{coveringpi}
An optimal policy $\pi_{x}$ is \textbf{covering} if $O_{\mx}(s_{x}, a_{x}) = 1 \Rightarrow a_{x} \in \text{supp}(\pi_{x}(\cdot | s_{x}))$.
\label{def:coveringpi}
\end{restatable}

\begin{restatable}{definition}{unichain}
MDP $\mx$ is \textbf{unichain}, if all policies induce irreducible Markov Chains and all stochastic optimal policies induce ergodic, i.e irreducible and aperiodic, Markov Chains.  
\label{def:unichain}
\end{restatable}

\begin{restatable}{assumption}{regularity} 
\label{ass:regularity}
All considered MDPs are unichain with discrete state, action spaces and deterministic dynamics i.e. $P: \mc{S} \times \mc{A} \rightarrow \mc{S}$. Furthermore, there exists dummy state, actions $s^{d}, a^{d}$ where $O_{\mc{M}}(s, a^{d}) = 0 ~ \forall s \in \S$ and $O_{\mc{M}}(s^{d}, a) = 0 ~ \forall a \in \A$
\end{restatable}

As stated in Assumption \ref{ass:regularity}, we consider discrete unichain MDPs with deterministic dynamics. This assumption is weak since physics is largely deterministic and many control behaviors, such as walking, are described by unichains.

\subsection{Proof of Theorem \ref{lemma:pi_adapt}}
\label{sec:app:proofs_t1}
\piadapt*
\begin{proof}
Without loss of generality, consider an arbitrarily chosen sample $a_{x} = g(a_{y}), a_{y} \sim \pi_{y}(\cdot | \phi(s_{x}))$ for any $s_{x} \in \sx$. We first see that: 
\begin{align}
O_{\my}\big(\phi(s_{x}), \psi(a_{x}) \big) = O_{\my}\big(\phi(s_{x}), \psi(g(a_{y}))\big) = O_{\my}\big(\phi(s_{x}), a_{y}\big) = 1
\end{align}
where the first step substitutes $a_{x} = g(a_{y})$, the second step applies $\psi \circ g(a_{y}) = a_{y}$ since $O(\phi(s_{x}), a_{y}) = 1$ due to the optimality of $\pi_{y}$, and the last step follows from Corollary \ref{cor:optind}. Since $(\phi, \psi)$ is a reduction, we have that $O_{\my}(\phi(s_{x}), \psi(a_{x})) = 1 \Rightarrow O_{\mx}(s_{x}, a_{x}) = 1$ by Equation (1). Therefore, $O_{\mx}(s_{x}, a_{x}) = 1 ~~ \forall s_{x} \in \sx, \forall a_{x} \in \text{supp}(\hat{\pi}_{x}(\cdot | s_{x}))$. Then by Lemma \ref{lemma:mixturepi}, $\hat{\pi}_{x}$ is optimal.
\end{proof}

\subsection{Proof of Theorem \ref{thm:main}}
We first introduce some lemmas necessary to proving our main theorem. 
\begin{restatable}{lemma}{mixturepi}
Let MDP $\mx$ satisfy Assumption \ref{ass:regularity} and $\pi_{x}(a_{x} | s_{x})$ be a (stochastic) mixture policy that chooses $a_{x}$ randomly from $\{a_{x} | O(s_{x}, a_{x}) = 1\}$. Then, $\pi_{x}$ is optimal. \citep{ortner2018mixture}
\label{lemma:mixturepi}
\end{restatable}

\begin{corollary}\label{cor:optind}
Let MDP $\mx$ satisfy Assumption \ref{ass:regularity} and $\pi_{x}$ be optimal. Then $O_{\mx}(s_{x}, a_{x}) = 1 ~ \forall s_{x} \in \sx, a_{x} \in \text{supp}(\pi_{x}(\cdot | s_{x}))$
\end{corollary}

\begin{restatable}{lemma}{tripletstationary}
\label{lem:stationary-dist}
Let MDP $\mx$ satisfy Assumption \ref{ass:regularity} and $\pi_{x}$ be a stochastic optimal policy. Then the triplet stationary distribution ${\rho}_{\pi_{x}}^{x}(s_{x}, a_{x}, s_{x}') = \lim_{t \to \infty} \Pr(s_{x}^{(t)} = s_{x}, a_{x}^{(t)} = a_{x}, s_{x}^{(t+1)} = s_{x}'; \pi_{x}, \px, \eta_{x})$
exists and is unique.
\end{restatable}
\begin{proof}
\begin{align*}
{\rho}_{\pi_{x}}^{x}(s_{x}, a_{x}, s_{x}') &= \lim_{t \to \infty} \Pr(s_{x}^{(t)} = s_{x}, a_{x}^{(t)} = a_{x}, s_{x}^{(t+1)} = s_{x}'; \pi_{x}, \px, \eta_{x}) \\
&= \lim_{t \to \infty} \Pr(s_{x}^{(t)} = s_{x}; \pi_{x}, \px, \eta_{x}) \pi_{x}(a_{x} | s_{x}) \mathbbm{1}(s_{x}' = \px(s_{x}, a_{x})) \\
&= \pi_{x}(a_{x} | s_{x}) \mathbbm{1}(s_{x}' = \px(s_{x}, a_{x})) \lim_{t \to \infty} \Pr(s_{x}^{(t)} = s_{x}; \pi_{x}, \px, \eta_{y})
\end{align*}
where $\mathbbm{1}$ is the indicator function. The limit in the last line is the stationary distribution over states, which exists and is unique since a stochastic optimal policy induces an ergodic Markov Chain over states. 
\end{proof}

\begin{lemma}\label{lemma:limit-mean}
If a real sequence $\{a_i \}_{i=1}^{\infty}$ 
converges to some $a \in \bb{R}$, then 
$$
\lim_{T \to \infty} \frac{1}{T} \sum_{i=1}^{T} a_i = \lim_{i \to \infty} a_i = a
$$
\end{lemma}
\begin{proof}
Denote $A_T = \sum_{i=1}^{T} a_i$, and $B_T = T$. We have
\begin{align}
    \lim_{T \to \infty} \frac{A_{T+1} - A_{T}}{B_{T+1} - B_T} = \lim_{T \to \infty} a_{T+1} = a
\end{align}
According to the Stolz–Cesàro theorem, 
$$
\lim_{T \to \infty} \frac{A_{T+1} - A_{T}}{B_{T+1} - B_T} = \lim_{T \to \infty} \frac{A_{T}}{B_T}
$$
if the limit on the left hand side exists.
Therefore
\begin{align}
    \lim_{T \to \infty} \frac{A_{T}}{B_T} = \lim_{T \to \infty} \frac{1}{T} \sum_{i=1}^{T} a_i = a
\end{align}
which completes the proof.
\end{proof}

Recall that our target distribution $\sigyy$ and proxy distribution $\sigxy$ were defined as: 

\begin{align}
\sigyy(s_y, a_y, s'_y) &= \lim_{T \to \infty} \frac{1}{T} \sum_{t=0}^{T-1} \Pr(s^{(t)}_y = s_y, a^{(t)}_y = a_y, s^{(t+1)}_y = s'_y; \pi_{y}, \py, \eta_{y})\\
\sigxy(s_y, a_y, s'_y) &= \lim_{T \to \infty} \frac{1}{T} \sum_{t=0}^{T-1} \Pr(\hat{s}_y^{(t)} = {s}_y, \hat{a}_y^{(t)} = {a}_y, \hat{s}_y^{(t+1)} = {s}'_y; \gP)
\end{align}

We are now ready to prove that our proxy and target limiting distributions exist. 


\begin{restatable}{lemma}{limity}
\label{lem:limit_y}
Let MDP $\my$ satisfy Assumption \ref{ass:regularity} and $\pi_{y}$ be a stocahstic optimal policy. Then, $\sigyy(s_y, a_y, s'_y) = \rho_{\pi_y}^{y}(s_y, a_y, s'_y)$.
\end{restatable}
\begin{proof}
Recall that the stationary distribution $\rho_{\pi_y}^{y}(s_y, a_y, s'_y)$ is the following limiting distribution:
\begin{align}
{\rho}_{\pi_y}^{y}(s_{y}, a_{y}, s_{y}') = \lim_{t \to \infty} \Pr(s_{y}^{(t)} = s_{y}, a_{y}^{(t)} = a_{y}, s_{y}^{(t+1)} = s_{y}'; \pi_{y}, \py, \eta_{y})
\end{align}
$\rho_{\pi_y}^{y}(s_y, a_y, s'_y)$ exist for $\my$ as shown in Lemma \ref{lem:stationary-dist}. Then, 
\begin{align}
    \sigyy(s_y, a_y, s'_y) & = \lim_{T \to \infty}\frac{1}{T} \sum_{t=0}^{T-1} \Pr(s_{y}^{(t)} = s_{y}, a_{y}^{(t)} = a_{y}, s_{y}^{(t+1)} = s_{y}'; \pi_{y}, \py, \eta_{y}) \\
    & = \lim_{t \to \infty} \Pr(s_{y}^{(t)} = s_{y}, a_{y}^{(t)} = a_{y}, s_{y}^{(t+1)} = s_{y}'; \pi_{y}, \py, \eta_{y}) \\
    & = {\rho}_{\pi_y}^{y}(s_{y}, a_{y}, s_{y}')
\end{align}
as desired. The second line follows from Lemma \ref{lemma:limit-mean} and the last line follows from Lemma \ref{lem:stationary-dist}.
\end{proof}

\newpage

\begin{restatable}{lemma}{support_y}
\label{lemma:supp_y}
Let MDP $\my$ satisfy Assumption \ref{ass:regularity} and $\pi_{y}$ be a stochastic optimal policy. Then,  
\begin{align*}
\text{supp}(\sigyy) \subseteq \{(s_{y}, a_{y}, s_{y}') | O_{\my}(s_{y}, a_{y}) = 1, s_{y}, s_{y}' \in \sy, a_{y} \in \ay\}
\end{align*}
\end{restatable}
\begin{proof}
Assume for contradiction that there exists $(s_{y}, a_{y}, s_{y}') \in \text{supp}(\sigyy)$ but $(s_{y}, a_{y}, s_{y}') \notin \{(s_{y}, a_{y}, s_{y}') | O_{\my}(s_{y}, a_{y}) = 1, s_{y}, s_{y}' \in \sy, a_{y} \in \ay\}$. Then $O_{\my}(s_{y}, a_{y}) = 0$. Since 
\begin{align*}
    \sigyy(s_{y}, a_{y}, a_{y}) &=  \lim_{t \to \infty} \Pr(s_{y}^{(t)} = s_{y}, a_{y}^{(t)} = a_{y}, s_{y}^{(t+1)} = s_{y}'; \pi_{y}, \py, \eta_{y}) \\
     &= \lim_{t \to \infty} \Pr(s_{y}^{(t)} = s_{y}) \Pr(a_{y}^{(t)} = a_{y} | s_{y}^{(t)} = s_{y}) \Pr(s_{y}^{(t+1)} = s_{y}' | s_{y}^{(t)} = s_{y}, a_{y}^{(t)} = a_{y}) \\ 
     &= \lim_{t \to \infty} \Pr(s_{y}^{(t)} = s_{y}) \pi_{y}(a_{y} | s_{y})\Pr(s_{y}^{(t+1)} = s_{y}' | s_{y}^{(t)} = s_{y}, a_{y}^{(t)} = a_{y}) \\
     &= \cancelto{0}{\pi_{y}(a_{y} | s_{y})} \Pr(s_{y}^{(t+1)} = s_{y}' | s_{y}^{(t)} = s_{y}, a_{y}^{(t)} = a_{y}) \lim_{t \to \infty} \Pr(s_{y}^{(t)} = s_{y}) \\
     &= 0
\end{align*}
First line follows from Lemma \ref{lem:limit_y} and terms are taken out of the limit in the second to last line since the stationary distribution over states exist as $\my$ is unichain and $\pi_{y}$ is stochastic optimal. $\pi_{y}(a_{y} | s_{y}) = 0$ since $O_{\my}(s_{y}, a_{y}) = 0 \Rightarrow \pi_{y}(a_{y} | s_{y}) = 0$ from Corollary \ref{cor:optind}. Then, we have $\sigyy(s_{y}, a_{y}, a_{y}) = 0$ which contradicts $(s_{y}, a_{y}, s_{y}') \in \text{supp}(\sigyy)$ concluding the proof. 
\end{proof}


\begin{restatable}{lemma}{limitx}
\label{lem:limit_x}
Let MDP $\mx$ satisfy Assumption \ref{ass:regularity} and $\hat{\pi}_{x} = g \circ \pi_{y} \circ f$ be an stochastic optimal policy in $\mx$ where $f: \sx \rightarrow \sy$ is the state map, $g: \ay \rightarrow \ax$ is injective action map, and $\pi_{y}$ is a stochastic optimal policy in $\my$. Further let $\mc{F}: \sx \times g(\ay) \times \sx \rightarrow \sy \times \ay \times \sy$ be the map $\mc{F}(a, b, c) = (f(a), g^{-1}(b), f(c))$. Then, $\sigxy(s_y, a_y, s'_y) = \mc{F}(\rho_{\hat{\pi}_{x}}^{x}(s_x, a_x, s'_x))$. 
\end{restatable}

\begin{proof}
We first define the triplet random variables $X^{(t)} = (s_x^{(t)}, a_x^{(t)}, s_x^{(t+1)})$ for $t = 0, 1, 2, ...$ where $s_{x}^{(t)}, a_{x}^{(t)}, s_x^{(t+1)}$ for $t = 0, 1, 2, ...$ were defined in Definition \ref{def:co-domain-exec}. $\mc{F}$ is a function on $supp(\rhoxx) \in \sx \times g(\ay) \times \sx$ and $\mc{F}(X^{(t)}) = (\hat{s}_y^{(t)}, \hat{a}_y^{(t)}, \hat{s}_y^{(t+1)})$. Furthermore, since $\mc{F}$ is a function defined on a discrete domain and codomain, there always exists a trivial continuous extension of $\mc{F}$. We may thus apply the continuous mapping theorem \citep{billingsley2014convergence}: 
$$
X^{(t)} \xrightarrow{d} X \Rightarrow \mc{F}(X^{(t)}) \xrightarrow{d} \mc{F}(X)
$$
Since $\mx$ is unichain and $\hat{\pi}_{x}$ is stochastic optimal, the distribution of $X^{(t)}$ converges (in distribution) to $\rho_{\hat{\pi}_{x}}^x(s_x, a_x, s'_x)$ as $t \rightarrow \infty$ by Lemma \ref{lem:stationary-dist}. Applying the continuous mapping theorem, it follows that the distribution of $\mc{F}(X^{(t)}) = (\hat{s}_y^{(t)}, \hat{a}_y^{(t)}, \hat{s}_y^{(t+1)})$ converges (in distribution) to the pushforward measure $\mc{F}(\rho_{\hat{\pi}_{x}}^x(s_x, a_x, s'_x))$ as $t \rightarrow \infty$

Directly applying this result, we obtain:
\begin{align}
\sigxy({s}_y, {a}_y, {s}'_y) &= \lim_{T \to \infty} \frac{1}{T} \sum_{t=0}^{T-1} \Pr(\hat{s}_y^{(t)} = {s}_y, \hat{a}_y^{(t)} = {a}_y, \hat{s}_y^{(t+1)} = {s}'_y; \gP) \label{eq:limiting} \\
&= \lim_{t \to \infty} \Pr(\hat{s}_y^{(t)} = {s}_y, \hat{a}_y^{(t)} = {a}_y, \hat{s}_y^{(t+1)} = {s}'_y; \gP) \label{eq:stationary} \\
&= \mc{F}(\rho_{\hat{\pi}_{x}}^x(s_x, a_x, s'_x)) \label{eq:mapend}
\end{align}
as desired. Line (\ref{eq:limiting}) $\rightarrow$ (\ref{eq:stationary}) follows from Lemma \ref{lemma:limit-mean} and (\ref{eq:stationary}) $\rightarrow$ (\ref{eq:mapend}) follows from the continuous mapping theorem. 
\end{proof}

\begin{restatable}{lemma}{indicator}
\label{lemma:indicator}
Let $X, Y$ be countable sets, $\phi: X \rightarrow Y$ be a function, and $\mathbbm{1}$ be the indicator function. We denote $\phi^{-1}(y) = \{x | \phi(x) = y\}$. Then $\forall x \in X, y \in Y$
\begin{align*}
\mathbbm{1}(y = \phi(x)) = \sum_{z \in \phi^{-1}(y)} \mathbbm{1}(x = z)
\end{align*}
\end{restatable}
\begin{proof}
Since both the left and right hand-side of the desired equality only take on values in $\{0, 1\}$, it suffices to show the following statements hold for arbitrarily chosen $x \in X, y \in Y$:
\begin{align*}
\sum_{z \in \phi^{-1}(y)} \mathbbm{1}(x = z) = 1 &\Rightarrow \mathbbm{1}(y = \phi(x)) = 1 \\
\mathbbm{1}(y = \phi(x)) = 1 &\Rightarrow \sum_{z \in \phi^{-1}(y)} \mathbbm{1}(x = z) = 1
\end{align*}
For the first direction, we see that if $\sum_{z \in \phi^{-1}(y)} \mathbbm{1}(x = z) = 1$, then $x \in \phi^{-1}(y)$, and thus $\phi(x) = y$. 

For the second direction if $\mathbbm{1}(y = \phi(x)) = 1$, then $x \in \phi^{-1}(y)$. Thus there exists a unique $z$ such that $z = x$ and $z \in \phi^{-1}(y)$. Then, $\sum_{z \in \phi^{-1}(y)} \mathbbm{1}(x = z) = 1$ as desired, which concludes the proof. 
\end{proof}

Finally, we prove the main theorem. Recall that the optimization objectives are: 1. optimality of $\hat{\pi}_{x}$ 2. $\sigxy = \sigyy$. 

\main*
\begin{proof}
We first show the ($\Rightarrow$) direction. Using any $(\phi, \psi) \in \Gamma(\gM_x, \gM_y)$ we construct $f$ and $g$ in the following manner: $f(s_x) = \phi(s_x) ~~ \forall s_x \in \sx$. $g(a_{y})$ maps to an arbitrary chosen element from the set $\psi^{-1}(a_{y}) = \{a_{x} | \psi(a_{x}) = a_{y}\}$ if $\psi^{-1}(a_{y}) \neq \emptyset$ and an arbitrarily chosen action $a_{x} \in \ax$ otherwise. We see that $\forall a_{y} \in \ay$ for which $\exists s_{y} \in \sy$ such that $O_{\my}(s_{y}, a_{y}) = 1$, it holds that $\psi^{-1}(a_{y}) \neq \emptyset$ by Eq \ref{eq:surjective}.
Therefore, $\psi \circ g(a_{y}) = a_{y} ~~ \forall a_{y} \in \ay$ for which $\exists s_{y}$ such that $O_{\my}(s_{y}, a_{y}) = 1$ since $\psi$ maps all elements in $\psi^{-1}(a_{y})$ to $a_{y}$. For $\pi_{y}$ we choose any covering optimal policy for $\my$. It suffices to show that this choice of $f, g, \pi_{y}$ satisfies objectives 1, 2. 

$\bullet$ Objective 1. $\hat{\pi}_{x}$ is optimal: follows from Lemma \ref{lemma:pi_adapt}. 

$\bullet$ Objective 2. $\sigxy = \sigyy$: Since $f = \phi$ is a reduction, it follows that $\forall s_{y} \in \sy, a_{y} \in \ay$ such that $O_{\my}(s_{y}, a_{y}) = 1$, any $s_{y}' \in \sy$, and $\forall t = 0, 1, 2, ...$: 

\resizebox{\linewidth}{!}{
\begin{minipage}{\linewidth}
{\thickmuskip=0mu
\begin{align}
&\Pr(\hat{s}_y^{(t+1)}=s'_y | \hat{s}_y^{(t)}=s_y, \hat{a}_{y}^{(t)}=a_{y}) \nonumber \\
\qquad &= \sum\limits_{s_{x}'\in \sx} \Pr(\hat{s}_{y}^{(t+1)}=s_{y}' | \hat{s}_{x}^{(t+1)}=s_{x}', \hat{s}_y^{(t)}=s_y, \hat{a}_{y}^{(t)}=a_{y}) \Pr(\hat{s}_{x}^{(t + 1)}=s_{x}' | \hat{s}_y^{(t)}=s_y, \hat{a}_{y}^{(t)}=a_{y}) \nonumber \\
&= \sum\limits_{s_{x}'\in\sx} \Pr(\hat{s}_{y}^{(t+1)}=s_{y}' | \hat{s}_{x}^{(t+1)}=s_{x}') \sum\limits_{\substack{s_{x} \in \sx \\ a_{x} \in \ax}} \Pr(\hat{s}_{x}^{(t + 1)}=s_{x}' | s_{x}^{(t)}=s_{x}, a_{x}^{(t)}=a_{x}, \hat{s}_y^{(t)}=s_y, \hat{a}_{y}^{(t)}=a_{y}) 
\Pr(s_{x}^{(t)}=s_{x}, a_{x}^{(t)}=a_{x} | \hat{s}_y^{(t)}=s_y, \hat{a}_{y}^{(t)}=a_{y}) \nonumber\\
&= \sum\limits_{s_{x}'\in\sx} \mathbbm{1}(s_{y}' = \phi(s_{x}')) \sum\limits_{\substack{s_{x} \in \sx \\ a_{x} \in \ax}} \Pr(\hat{s}_{x}^{(t + 1)}=s_{x}' | s_{x}^{(t)}=s_{x}, a_{x}^{(t)}=a_{x}) 
\Pr(s_{x}^{(t)}=s_{x} | a_{x}^{(t)}=a_{x}, \hat{s}_y^{(t)}=s_y, \hat{a}_{y}^{(t)}=a_{y}) \Pr(a_{x}^{(t)}=a_{x} | \hat{s}_y^{(t)}=s_y, \hat{a}_{y}^{(t)}=a_{y}) \nonumber\\
&= \sum\limits_{s_{x}'\in\sx} \mathbbm{1}(s_{y}' = \phi(s_{x}')) \sum\limits_{\substack{s_{x} \in \sx \\ a_{x} \in \ax}} \mathbbm{1}(s_{x}' = \px(s_{x}, a_{x})) 
\Pr(s_{x}^{(t)}=s_{x} | \hat{s}_y^{(t)}=s_y) \Pr(a_{x}^{(t)}=a_{x} | \hat{a}_{y}^{(t)}=a_{y}) \nonumber\\
&= \sum\limits_{s_{x}'\in\sx} \mathbbm{1}(s_{y}' = \phi(s_{x}')) \sum\limits_{\substack{s_{x} \in \sx \\ a_{x} \in \ax}} \mathbbm{1}(s_{x}' = \px(s_{x}, a_{x})) 
\dfrac{\Pr(\hat{s}_y^{(t)}=s_y | s_{x}^{(t)}=s_{x}) \Pr(s_{x}^{(t)}=s_{x})}{\sum \limits_{s_{x}'' \in \sx} \Pr(\hat{s}_y^{(t)}=s_y | s_{x}^{(t)}=s_{x}'') \Pr(s_{x}^{(t)}=s_{x}'') } \mathbbm{1}(a_{x} = g(a_{y})) \nonumber\\
&= \sum\limits_{s_{x}'\in\sx} \mathbbm{1}(s_{y}' = \phi(s_{x}')) \sum\limits_{s_{x} \in \sx} \mathbbm{1}(s_{x}' = \px(s_{x}, g(a_{y}))) 
\dfrac{\mathbbm{1}(s_y = \phi(s_{x})) \Pr(s_{x}^{(t)}=s_{x})}{\sum \limits_{s_{x}'' \in \sx} \mathbbm{1}(s_y = \phi(s_{x}'')) \Pr(s_{x}^{(t)}=s_{x}'')} \nonumber\\
&= \sum\limits_{s_{x}' \in \phi^{-1}(s_{y}')} \sum\limits_{s_{x} \in \phi^{-1}(s_{y})} \mathbbm{1}(s_{x}' = \px(s_{x}, g(a_{y}))) 
\dfrac{\Pr(s_{x}^{(t)}=s_{x})}{\sum \limits_{s_{x}'' \in \phi^{-1}(s_{y})} \Pr(s_{x}^{(t)}=s_{x}'')} \nonumber\\
&= \sum\limits_{s_{x} \in \phi^{-1}(s_{y})} \dfrac{\Pr(s_{x}^{(t)}=s_{x})}{\sum \limits_{s_{x}'' \in \phi^{-1}(s_{y})} \Pr(s_{x}^{(t)}=s_{x}'')} \sum\limits_{s_{x}' \in \phi^{-1}(s_{y}')} \mathbbm{1}(s_{x}' = \px(s_{x}, g(a_{y}))) \nonumber\\
&\stackrel{\text{Lemma} \ref{lemma:indicator}}{=} \sum\limits_{s_{x} \in \phi^{-1}(s_{y})} \dfrac{\Pr(s_{x}^{(t)}=s_{x})}{\sum \limits_{s_{x}'' \in \phi^{-1}(s_{y})} \Pr(s_{x}^{(t)}=s_{x}'')} \mathbbm{1}\Big(s_{y}' = \phi \Big(\px \big(s_{x}, g(a_{y}) \big) \Big)\Big) \nonumber\\
&\stackrel{\text{Eq} \ref{eq:transition}}{=} \sum\limits_{s_{x} \in \phi^{-1}(s_{y})} \dfrac{\Pr(s_{x}^{(t)}=s_{x})}{\sum \limits_{s_{x}'' \in \phi^{-1}(s_{y})} \Pr(s_{x}^{(t)}=s_{x}'')} \mathbbm{1}(s_{y}' = \py(s_{y}, a_{y})) \nonumber\\
&= \mathbbm{1}(s_{y}' = \py(s_{y}, a_{y})) ~ = ~ \Pr(s_y^{(t+1)}=s'_y | s_y^{(t)}=s_y, a_{y}^{(t)}=a_{y}) \label{eq:mt_1}
\end{align}}
\end{minipage}
}
Furthermore, from Definition \ref{def:co-domain-exec}, we have: 
\begin{align}
\Pr(\hat{a}_{y}^{(t)} = a_{y} | \hat{s}_y^{(t)} = s_y) &= \pi_y(a_{y} | s_{y}) = \Pr(a_{y}^{(t)} = a_{y} | s_y^{(t)} = s_y) \label{eq:mt_2}
\end{align}
Then, $\forall s_{y}, s_{y}' \in \sy$ and $\forall t = 0, 1, 2, ...$
\begin{align}
\Pr(\hat{s}_{y}^{(t+1)} = s_{y} | \hat{s}_y^{(t)} = s_y) &= \sum\limits_{a_{y} \in \ay} \Pr(\hat{s}_{y}^{(t+1)} = s_{y} | \hat{s}_y^{(t)} = s_y, \hat{a}_{y}^{(t)} = a_{y}) \Pr(\hat{a}_{y}^{(t)} = a_{y} | \hat{s}_y^{(t)} = s_y) \nonumber\\
&= \sum\limits_{ a_{y} \in \text{supp}(\pi_{y}(\cdot | s_{y})) } \Pr(s_{y}^{(t+1)} = s_{y} | s_y^{(t)} = s_y, a_{y}^{(t)} = a_{y}) \pi_{y}(a_{y} | s_{y}) \nonumber\\
&= \Pr(s_{y}^{(t+1)} = s_{y} | s_y^{(t)} = s_y) 
\end{align}
we are justified in the substitution for the dynamics in the second line since $O_{\my}(s_{y}, a_{y}) = 1 \forall s_{y} \in \sy, a_{y} \in \text{supp}(\pi_{y}(\cdot | s_{y}))$ by Corollary \ref{cor:optind}. Since $\my$ is unichain and $\pi_{y}$ is a stochastic optimal policy, the stationary distribution $\lim_{t \rightarrow \infty} \Pr(s_{y}^{(t)} = s_{y})$ is invariant to the initial state distribution $\eta_{y}$ and only depends on the state transition dynamics $\Pr(s_y^{(t+1)} = s'_y | s_y^{(t)} = s_y)$. Equivalently any stochastic process with the same state transition dynamics will converge to the same stationary distribution regardless of the initial state distribution. Thus,  
\begin{align}
\lim_{t \rightarrow \infty} \Pr(\hat{s}_{y}^{(t)} = s_{y}) = \lim_{t \rightarrow \infty} \Pr(s_{y}^{(t)} = s_{y}) \quad \forall s_{y} \in \sy \label{eq:mt_3}
\end{align}

Finally putting these results together, the following equalities hold for $(s_{y}, a_{y}, s_{y}') \in \{(s_{y}, a_{y}, s_{y}') | O_{\my}(s_{y}, a_{y}) = 1, s_{y}, s_{y}' \in \sy, a_{y} \in \ay\}$
\begin{align*}
\sigxy(s_{y}, a_{y}, s_{y}') \stackEq{\text{Lemma} \ref{lem:limit_y}} \lim_{t \to \infty} \Pr(\hat{s}_y^{(t)} = {s}_y, \hat{a}_y^{(t)} = {a}_y, \hat{s}_y^{(t+1)} = {s}'_y; \gP) \\
\stackEq{} \lim_{t \to \infty} \Pr(\hat{s}_{y}^{(t)} = s_{y}) \Pr(\hat{a}_{y}^{(t)}=a_{y} | \hat{s}_{y}^{(t)}=s_{y}) \Pr(\hat{s}_y^{(t+1)} = s_y' | \hat{s}_y^{(t)} = s_y, \hat{a}_y^{(t)} = a_y) \\
\stackEq{\text{Eq} (\ref{eq:mt_1}),(\ref{eq:mt_2})} \lim_{t \to \infty} \Pr(\hat{s}_{y}^{(t)} = s_{y}) \Pr(a_{y}^{(t)}=a_{y} | s_{y}^{(t)}=s_{y}) \Pr(s_y^{(t+1)} = s_y' | s_y^{(t)} = s_y, a_y^{(t)} = a_y) \\
\stackEq{} \pi_{y}(a_{y} | s_{y}) \mathbbm{1}(s_{y}' = \py(s_{y}, a_{y})) \lim_{t \to \infty} \Pr(\hat{s}_{y}^{(t)} = s_{y}) \\
\stackEq{\text{Eq} (\ref{eq:mt_3})} \pi_{y}(a_{y} | s_{y}) \mathbbm{1}(s_{y}' = \py(s_{y}, a_{y})) \lim_{t \to \infty} \Pr(s_{y}^{(t)} = s_{y}) \\
\stackEq{} \lim_{t \to \infty} \Pr(s_{y}^{(t)} = s_{y}) \Pr(a_{y}^{(t)}=a_{y} | s_{y}^{(t)}=s_{y}) \Pr(s_y^{(t+1)} = s_y' | s_y^{(t)} = s_y, a_y^{(t)} = a_y) \\
\stackEq{\text{Lemma} \ref{lem:limit_y}} \sigyy(s_{y}, a_{y}, s_{y}')
\end{align*}
The constant terms are moved in and out of the limit since the stationary distribution over states exist as $\my$ is unichain and $\pi_{y}$ is optimal in $\my$. This allows us to conclude that $\sigxy = \sigyy$ since $\sigyy$ is supported on $\{(s_{y}, a_{y}, s_{y}') | O_{\my}(s_{y}, a_{y}) = 1, s_{y}, s_{y}' \in \sy, a_{y} \in \ay\}$ by Lemma \ref{lemma:supp_y}. 

Now we show the ($\Leftarrow$) direction. We first introduce some overloaded notation: 
\begin{align}
    \sigxx(s_{x}) &= \lim_{T \to \infty}\frac{1}{T} \sum_{t=0}^{T-1} \Pr(s_{x}^{(t)} = s_{x}; \hat{\pi}_{x}, \px, \eta_{x}) 
    \stackrel{\text{Lemma} \ref{lem:limit_y}}{=} \lim_{t \to \infty}\Pr(s_{x}^{(t)} = s_{x}; \hat{\pi}_{x}, \px, \eta_{x}) \nonumber\\
    \sigxx(s_{x}, a_{x}) \stackEq{} \lim_{T \to \infty}\frac{1}{T} \sum_{t=0}^{T-1} \Pr(s_{x}^{(t)} = s_{x}, a_{x}^{(t)} = a_{x}; \hat{\pi}_{x}, \px, \eta_{x})\nonumber\\
    \stackEq{\text{Lemma} \ref{lem:limit_y}} \lim_{t \to \infty} \Pr(s_{x}^{(t)} = s_{x}, a_{x}^{(t)} = a_{x}; \hat{\pi}_{x}, \px, \eta_{x})\nonumber \\
    \stackEq{} \lim_{t \to \infty} \Pr(s_{x}^{(t)} = s_{x}; \hat{\pi}_{x}, \px, \eta_{x}) \hat{\pi}_{x}(a_{x} | s_{x})\nonumber \\
    \stackEq{} \hat{\pi}_{x}(a_{x} | s_{x}) \lim_{t \to \infty} \Pr(s_{x}^{(t)} = s_{x}; \hat{\pi}_{x}, \px, \eta_{x}) \nonumber \\
    \stackEq{} \hat{\pi}_{x}(a_{x} | s_{x}) \sigxx(s_{x}) \label{eq:sig1tosig2}
\end{align}
Then, 
\begin{align}
\sigxx(s_{x}, a_{x}, s_{x}') \stackEq{\text{Lemma} \ref{lem:limit_y}} \lim_{t \to \infty} \Pr(s_{x}^{(t)} = s_{x}, a_{x}^{(t)} = a_{x}, s_{x}^{(t+1)} = s_{x}'; \hat{\pi}_{x}, \px, \eta_{x}) \nonumber \\
\stackEq{} \lim_{t \to \infty} \Pr(s_{x}^{(t+1)} = s_{x}' | s_{x}^{(t)} = s_{x}, a_{x}^{(t)} = a_{x}) \Pr(s_{x}^{(t)} = s_{x}, a_{x}^{(t)} = a_{x}; \hat{\pi}_{x}, \px, \eta_{x}) \nonumber \\
\stackEq{} \mathbbm{1}(s_{x}' = \px(s_{x}, a_{x})) \lim_{t \to \infty} \Pr(s_{x}^{(t)} = s_{x}, a_{x}^{(t)} = a_{x}; \hat{\pi}_{x}, \px, \eta_{x}) \nonumber\\
\stackEq{} \mathbbm{1}(s_{x}' = \px(s_{x}, a_{x})) \sigxx(s_{x}, a_{x}) \label{eq:sig2tosig3} \\
\stackEq{} \mathbbm{1}(s_{x}' = \px(s_{x}, a_{x})) \hat{\pi}_{x}(a_{x} | s_{x}) \sigxx(s_{x}) \label{eq:sig1tosig3}
\end{align}
Given $f, g$ that satisfy objective 1, 2, and 3 we construct $(\phi, \psi)$ as follows and show that $(\phi, \psi) \in \Gamma(\gM_x, \gM_y)$:
\begin{align*}
\phi(s_{x}) &=
  \begin{cases}
       f(s_{x}) & \text{if $s_{x} \in \text{supp}(\sigxx(s_{x}))$} \\
       s_{y}^{d} & \text{otherwise} 
  \end{cases}
\\
\psi(a_{x}) &=
  \begin{cases}
       g^{-1}(a_{x}) & \text{if $a_{x} \in \A_{\hat{\pi}_{x}} = \bigcup_{s_{x} \in \sx} \text{supp}(\hat{\pi}_{x}(\cdot | s_{x}))$} \\
       a_{y}^{d} & \text{otherwise} 
  \end{cases}
\end{align*}

where $s_{y}^{d}, a_{y}^{d}$ are dummy state, actions such that $O_{\my}(s_{y}^{d}, a_{y}) = 0 ~~ \forall a_{y} \in \ay$ and $O_{\my}(s_{y}, a_{y}^{d}) = 0 ~~ \forall s_{y} \in \sy$. Such dummy state, actions always exist per Assumption \ref{ass:regularity}. Mapping to the dummy state, action will ensure that the constructions will not map suboptimal state, action pairs from domain $x$ to optimal state action pairs in domain $y$. The following statement holds for our construction $(\phi, \psi)$: 
\begin{align}
(s_{x}^*, a_{x}^*) \in \text{supp}(\sigxx(s_{x}, a_{x})) \iff O_{\my}(\phi(s_{x}^*), \psi(a_{x}^*)) = 1 \quad \forall s_{x}^* \in \sx, a_{x}^* \in \ax \label{eq:mt_subclaim}
\end{align}

We first prove the forward direction: $(s^*_{x}, a^*_{x}) \in \text{supp}(\sigxx(s_{x}, a_{x})) \Rightarrow \sigxx(s_{x}^*, a_{x}^*) \stackrel{\text{Eq} \ref{eq:sig1tosig2}}{=} \sigxx(s_{x})\hat{\pi}_{x}(a_{x}^* | s_{x}^*) > 0$, so $\sigxx(s_{x}^*) > 0$, i.e $s_{x}^* \in \text{supp}(\sigxx(s_{x}))$, and $\hat{\pi}_{x}(a_{x}^* | s_{x}^*) > 0$. Furthermore, $\hat{\pi}_{x}(a_{x}^* | s_{x}^*) > 0 \Rightarrow g^{-1}(a_{x}^*) \in \text{supp}(\pi_{y}(\cdot | f(s_{x}^*)))$ since $g$ is injective. To see this, assume $\exists (s_{x}^*, a_{x}^*)$ such that $\hat{\pi}_{x}(a_{x}^* | s_{x}^*) > 0$ but $g^{-1}(a_{x}^*) \notin \text{supp}(\pi_{y}(\cdot | f(s_{x}^*)))$. Then there must exists $a_{y}' \in \text{supp}(\pi_{y}(\cdot | f(s_{x}^*)))$ such that $a_{y}' \neq g^{-1}(a_{x}^*)$ but $g(a_{y}') = g(g^{-1}(a_{x}^*)) = a_{x}^*$ contradicting the injectivity of $g$ on $\ay$. Putting these results together we obtain $\psi(a_{x}^*) = g^{-1}(a_{x}^*) \in \text{supp}(\pi_{y}(\cdot | \phi(s_{x}^*)))$. Since $\pi_{y}$ is a stochastic optimal policy and $\my$ is unichain, $\psi(a_{x}^*) \in \text{supp}(\pi_{y}(\cdot | \phi(s_{x}^*))) \Rightarrow O_{\my}(\phi(s_{x}^*), \psi(a_{x}^*)) = 1$ by Corollary \ref{cor:optind}. 

For the converse direction we prove the contrapostive: $(s_{x}^*, a_{x}^*) \notin \text{supp}(\sigxx(s_{x}, a_{x})) \Rightarrow O_{\my}(\phi(s_{x}^*), \psi(a_{x}^*)) = 0 ~~ \forall s_{x} \in \sx, a_{x} \in \ax$. We exhaustively consider all cases in which $(s_{x}^*, a_{x}^*) \notin \text{supp}(\sigxx(s_{x}, a_{x}))$, i.e $\sigxx(s_{x}^*, a_{x}^*) \stackrel{\text{Eq} \ref{eq:sig1tosig2}}{=} \hat{\pi}_{x}(a_{x}^* | s_{x}^*) \sigxx(s_{x}^*) = 0$. 
If $\sigxx(s_{x}^*) = 0$, then $s_{x}^* \notin \text{supp}(\sigxx(s_{x}))$, so $O_{\my}(\phi(s_{x}^*), a_{y}) = O_{\my}(s_{y}^{d}, a_{y}) = 0 ~~ \forall a_{y} \in \ay$. 
Else if $\hat{\pi}_{x}(a_{x}^* | s_{x}^*) = 0, \sigxx(s_{x}^*) > 0$ and $a_{x} \notin \A_{\hat{\pi}_{x}}$ then $O_{\my}(s_{y}, \psi(a_{x}^*)) = O_{\my}(s_{y}, a_{y}^{d}) = 0 ~~ \forall s_{y} \in \sy$. 
Finally, consider the case $\hat{\pi}_{x}(a_{x}^* | s_{x}^*) = 0, \sigxx(s_{x}^*) > 0$ and $a_{x}^* \in \A_{\hat{\pi}_{x}}$. Assume for contradiction that $O_{\my}(\phi(s_{x}^*), \psi(a_{x}^*)) = 1$. Then, $\psi(a_{x}^*) \in \text{supp}(\pi_{y}(\cdot | \phi(s_{x}^*))$ since $\pi_{y}$ is a covering optimal policy from Definition \ref{def:coveringpi}, which implies $g^{-1}(a_{x}^*) \in \text{supp}(\pi_{y}(\cdot | f(s_{x}^*))$ since $\sigxx(s_{x}^*) > 0$ and $a_{x}^* \in \A_{\hat{\pi}_{x}}$. It follows that $g(g^{-1}(a_{x}^*)) \in \text{supp}(g(\pi_{y}(\cdot | f(s_{x}^*)))) \Rightarrow a_{x}^* \in \text{supp}(\hat{\pi}_{x}(\cdot | s_{x}^*))$ since $\hat{\pi}_{x}(\cdot | s_{x}^*)$ is the pushforward measure $g(\pi_{y}(\cdot | f(s_{x}^*)))$. Then, $\sigxx(s_{x}^*, a_{x}^*) \stackrel{\text{Eq} \ref{eq:sig1tosig2}}{=} \sigxx(s_{x}^*)\hat{\pi}_{x}(a_{x}^* | s_{x}^*) > 0$, since $\hat{\pi}_{x}(a_{x}^* | s_{x}^*) > 0$ and $\sigxx(s_{x}^*) > 0$, which contradicts $(s_{x}^*, a_{x}^*) \notin \text{supp}(\sigxx(s_{x}, a_{x}))$. This concludes the proof of Equation \ref{eq:mt_subclaim}. 

We proceed to show that the optimal policy and dynamics preservation properties hold for our construction $(\phi, \psi)$. 

$\bullet$ Optimality (Eq. \ref{eq:reward}): From the converse direction of the above subclaim and the optimality of $\hat{\pi}_{x}$ the result immediate follows: 
\begin{align*}
O_{\my}(\phi(s_{x}^*), \psi(a_{x}^*)) = 1 &\stackrel{\text{Eq} \ref{eq:mt_subclaim}}{\Rightarrow} (s_{x}^*, a_{x}^*) \in \text{supp}(\sigxx(s_{x}, a_{x})) \\
&\stackrel{\text{Eq} \ref{eq:sig1tosig2}}{\Rightarrow} \hat{\pi}_{x}(a_{x}^* | s_{x}^*) > 0 \\ &\stackrel{\text{Cor} \ref{cor:optind}}{\Rightarrow} O_{\mx}(s_{x}^*, a_{x}^*) = 1 ~~ \forall s_{x}^* \in \sx, a_{x}^* \in \ax
\end{align*}

$\bullet$ Surjection (Eq. \ref{eq:surjective}): Assume for contradiction $\exists (s_{y}^*, a_{y}^*)$ such that $O_{\my}(s_{y}^*, a_{y}^*) = 1$, but $\phi^{-1}(s_{y}^*) = \emptyset$ or $\psi^{-1}(a_{y}^*) = \emptyset$. Since $O_{\my}(s_{y}^*, a_{y}^*) = 1$ we have $s_{y}^* \neq s_{y}^{d}, a_{y}^* \neq a_{y}^{d}$. Thus $\phi(s_{y}^*)^{-1} = f^{-1}(s_{y}^*)$ and $\psi^{-1}(a_{y}^*) = \{(g^{-1})^{-1}(a_{y}^*)\} = \{g(a_{y}^*)\}$. Since $g$ is a function defined $\forall a_{y} \in \ay$, it follows that $\psi^{-1}(a_{y}^*) \neq \emptyset$. Thus it must be that $\phi^{-1}(s_{y}^*) = \emptyset$. Let $s_{y}^{*'} = \py(s_{y}^*, a_{y}^*)$. Then, 
\begin{align*}
\sigxy(s_{y}^*, a_{y}^*, s_{y}^{*'}) \stackEq{\text{Lemma} \ref{lem:limit_y}} \lim_{t \rightarrow \infty} \Pr(\hat{s}_{y}^{(t)} = s_{y}^*, \hat{a}_{y}^{(t)} = a_{y}^*, \hat{s}_{y}^{(t+1)} = s_{y}^{*'}) \\
\stackEq{} \lim_{t \rightarrow \infty} \Pr(\hat{s}_{y}^{(t+1)} = s_{y}^{*'} | \hat{s}_{y}^{(t)} = s_{y}^*, \hat{a}_{y}^{(t)} = a_{y}^*) \Pr(\hat{s}_{y}^{(t)} = s_{y}^*, \hat{a}_{y}^{(t)} = a_{y}^*) \\
\stackEq{} \lim_{t \rightarrow \infty} \Pr(\hat{s}_{y}^{(t+1)} = s_{y}^{*'} | \hat{s}_{y}^{(t)} = s_{y}^*, \hat{a}_{y}^{(t)} = a_{y}^*) \Pr(\hat{a}_{y}^{(t)} = a_{y}^* | \hat{s}_{y}^{(t)} = s_{y}^*) \Pr(\hat{s}_{y}^{(t)} = s_{y}^*) \\
\stackEq{} \lim_{t \rightarrow \infty} \Pr(\hat{s}_{y}^{(t+1)} = s_{y}^{*'} | \hat{s}_{y}^{(t)} = s_{y}^*, \hat{a}_{y}^{(t)} = a_{y}^*) \pi_{y}(a_{y}^* | s_{y}^*) \sum_{s_{x} \in \phi^{-1}(s_{y}^*)} \Pr(\hat{s}_{x}^{(t)} = s_{x}) \\
\stackEq{} \lim_{t \rightarrow \infty} \Pr(\hat{s}_{y}^{(t+1)} = s_{y}^{*'} | \hat{s}_{y}^{(t)} = s_{y}^*, \hat{a}_{y}^{(t)} = a_{y}^*) \pi_{y}(a_{y}^* | s_{y}^*) \cdot 0 \\
\stackEq{} 0
\end{align*}
However, 
\begin{align*}
\sigyy(s_{y}^*, a_{y}^*, s_{y}^{*'}) \stackEq{\text{Eq} \ref{eq:sig1tosig3}} \mathbbm{1}(\py(s_{y}^*, a_{y}^*) = \py(s_{y}^*, a_{y}^*)) \pi_{y}(a_{y}^* | s_{y}^*) \sigyy(s_{y}^*) \\
\stackEq{} \pi_{y}(a_{y}^* | s_{y}^*) \sigyy(s_{y}^*) > 0
\end{align*}
To see why the last inequality holds, first recall that $\my$ is unichain and $\pi_{y}$ is stochastic optimal for $\my$, so the stationary distribution over states have full support over $\sy$ ($\because$ stationary distributions of irreducible markov chains are fully supported over the entire state space) Therefore $\sigyy(s_{y}) \stackrel{\text{Lemma} \ref{lem:limit_y}}{=} \lim_{t \to \infty} \Pr(s_{y}^{(t)} = s_{y}; \pi_{y}, \py) > 0 \quad \forall s_{y} \in \sy$. Thus, we have $\sigyy(s_{y}^*) > 0$. Furthermore, $\pi_{y}(a_{y}^* | s_{y}^*) > 0$ by Corollary \ref{cor:optind}. Putting these two results together, we obtain $\sigyy(s_{y}^*, a_{y}^*, s_{y}^{*'}) > 0$. Then, $\sigxy \neq \sigyy$ which contradicts the satisfiability of objective 3. 

$\bullet$ Dynamics (Eq. \ref{eq:transition}): Assume for contradiction that $\exists s_{x}^{-}, a_{x}^{-}$ and $s_{x}^{-'} = \px(s_{x}^{-}, a_{x}^{-}$ such that $O_{\my}(\phi(s_{x}^{-}), \psi(a_{x}^{-})) = 1$ but the dynamics preservation property is violated, i.e $P_{y}(\phi(s_{x}^{-}), \psi(a_{x}^{-})) \neq \phi(P_{x}(s_{x}^{-}, a_{x}^{-})) = \phi(s_{x}^{-'})$. If $(s_{x}^{-}, a_{x}^{-}) \notin \text{supp}(\sigxx(s_{x}, a_{x}))$, then $O_{\my}(\phi(s_{x}^{-}), \psi(a_{x}^{-})) = 0$ by Equation \ref{eq:mt_subclaim} which contradicts $O(\phi(s_{x}^{-}), \psi(a_{x}^{-})) = 1$. Thus, it must be that $(s_{x}^{-}, a_{x}^{-}) \in \text{supp}(\sigxx(s_{x}, a_{x}))$ which further implies $(s_{x}^{-}, a_{x}^{-}, s_{x}^{-'}) \in \text{supp}(\sigxx(s_{x}, a_{x}, s_{x}'))$ by Equation \ref{eq:sig2tosig3} and $\phi(s_{x}^{-}) = f(s_{x}^{-}), \psi(a_{x}^{-}) = g^{-1}(a_{x}^{-})$ by Equation \ref{eq:sig1tosig2} since $\sigxx(s_{x}^{-}) > 0, \hat{\pi}(a_{x}^{-} | s_{x}^{-}) > 0$.

Let $\mc{F}: \sx \times g(\ay) \times \sx \rightarrow \sy \times \ay \times \sy$ be a function $(a, b, c) \mapsto (f(a), g^{-1}(b), f(c))$. Then, by Lemma \ref{lem:limit_x}, we have $\sigxy(s_{x}, a_{x}, s_{x}') = \mc{F}(\rhoxx(s_{x}, a_{x}, s_{x}')) \stackrel{\text{Lemma} \ref{lem:limit_y}}{=} \mc{F}(\sigxx(s_{x}, a_{x}, s_{x}'))$. So, 
\begin{align*}
\sigxx(s_{x}^{-}, a_{x}^{-}, s_{x}^{-'}) > 0 \Rightarrow \sigxy(\mc{F}(s_{x}^{-}, a_{x}^{-}, s_{x}^{-'})) = \sigxy(f(s_{x}^{-}), g^{-1}(a_{x}^{-}), f(s_{x}^{-'})) > 0 
\end{align*}

Thus, $(f(s_{x}^{-}), g^{-1}(a_{x}^{-}), f(s_{x}^{-'})) = (\phi(s_{x}^{-}), \psi(a_{x}^{-}), \phi(s_{x}^{-'})) \in \text{supp}(\sigxy(s_{x}, a_{x}, s_{x}'))$. However, 
\begin{align*}
\sigyy(\phi(s_{x}^{-}), \psi(a_{x}^{-}), \phi(s_{x}^{-'})) \stackEq{\text{Eq} \ref{eq:sig2tosig3}} \sigyy(\phi(s_{x}^{-}), \psi(a_{x}^{-})) \mathbbm{1}\big(\phi(s_{x}^{-'}) = \py(\phi(s_{x}^{-}), \psi(a_{x}^{-}))\big) \\
\stackEq{} \sigyy(\phi(s_{x}^{-}), \psi(a_{x}^{-})) \cdot 0 \\
\stackEq{} 0
\end{align*}
Thus, $\text{supp}(\sigxy) \neq \text{supp}(\sigyy) \Rightarrow \sigxy \neq \sigyy$ which contradicts $f, g$ satisfying objective 3. This concludes the proof of the main theorem.  
\end{proof}